\documentclass[letterpaper]{article} 
\usepackage{aaai25}  

\usepackage{times}  
\usepackage{helvet}  
\usepackage{courier}  
\usepackage{multirow}
\usepackage{amsmath}
\usepackage{tcolorbox}
\usepackage{xcolor, colortbl}
\definecolor{llmcolor}{RGB}{230,245,255}
\definecolor{humancolor}{RGB}{255,235,235}
\usepackage{bm}
\newcommand{\subfootnotesize}{\fontsize{8}{9}\selectfont}

\usepackage[hyphens]{url}  
\usepackage{graphicx} 
\usepackage{natbib}  
\usepackage{caption} 
\usepackage{algorithm}
\usepackage{algorithmic}

\usepackage{newfloat}
\usepackage{listings}
\DeclareCaptionStyle{ruled}{labelfont=normalfont,labelsep=colon,strut=off} 
\floatstyle{ruled}
\newfloat{listing}{tb}{lst}{}
\floatname{listing}{Listing}
\title{A Tale of Two Identities: An Ethical Audit of Human and AI-Crafted Personas}

\author{
    Pranav Narayanan Venkit, Jiayi Li\textsuperscript{\rm 1}, Yingfan Zhou\textsuperscript{\rm 1},\\
    Sarah Rajtmajer\textsuperscript{\rm 1}, Shomir Wilson\textsuperscript{\rm 1}
}
\affiliations{
    \textsuperscript{\rm 1}Pennsylvania State University, University Park, Pennsylvania, USA\\
    \{pranav.venkit, jpl6207, yxz5975, smr48, shomir\}@psu.edu

}

\usepackage{bibentry}

\begin{document}

\maketitle

\begin{abstract}
As LLMs (large language models) are increasingly used to generate synthetic personas—particularly in data-limited domains such as health, privacy, and HCI—it becomes necessary to understand how these narratives represent identity, especially that of minority communities. In this paper, we audit synthetic personas generated by 3 LLMs (GPT4o, Gemini 1.5 Pro, Deepseek v2.5) through the lens of representational harm, focusing specifically on racial identity. Using a mixed-methods approach combining close reading, lexical analysis, and a parameterized creativity framework, we compare 1,512 LLM-generated persona to human-authored responses. Our findings reveal that LLMs disproportionately foreground racial markers, overproduce culturally coded language, and construct personas that are syntactically elaborate yet narratively reductive. These patterns result in a range of sociotechnical harms—including stereotyping, exoticism, erasure, and benevolent bias—that are often obfuscated by superficially positive narrations. We formalize this phenomenon as algorithmic othering, where minoritized identities are rendered hypervisible but less authentic. Based on these findings, we offer design recommendations for narrative-aware evaluation metrics and community-centered validation protocols for synthetic identity generation.
\end{abstract}

%

\section{Introduction}

The expansion of LLMs as sociotechnical systems has transformed their usage in various spaces, including news and media, medicine and education \cite{sartori2022sociotechnical, narayanan2023towards}. Among their emerging uses is the automated generation of \textit{personas}—fictional but realistic representations of user identities. Originally developed within human-centered design to foster empathy and represent end-users in system development, personas are now frequently produced by LLMs at scale to support applications such as chatbot character design \cite{kim2023chatgpt, dam2024complete}, data augmentation \cite{whitehouse2023llm, ding2024data}, and even healthcare simulation and social science research \cite{bn2025thousand, biswas2023role, yukhymenko2024synthetic, manning2024automated, lehr2024chatgpt}.

Prior work has extensively documented biases in LLMs' classification \cite{ntoutsi2020bias, gupta2024sociodemographic} and generation behaviors \cite{ferrara2023fairness, ghosh2023chatgpt}, yet their representational harms in identity performance remain understudied. This gap is particularly concerning given personas' dual role as design tools and increasingly social actors \cite{lehr2024chatgpt, biswas2023role, barambones2024chatgpt}. Recent studies suggest LLM-generated text exhibit troubling patterns: overemphasizing trauma narratives for minority identities or flattening cultural complexity into reductive tropes \cite{cheng2023marked, lazik2025impostor, haxvig2024concerns}. These behaviors risk replicating historical harms in ethnographic research and sociotechnical cases, now automated at scale.

In this work, using the scope of representational harm identification and creativity analysis, we present a study of how LLMs construct identity through persona generation, with particular focus on representation within United States sociodemographic contexts. Through examination of 756 human-written self-descriptions (provided by 126 participants) alongside 1,512 LLM-generated personas from three leading models (GPT4o, Claude, and DeepSeek), we address three research questions:
\begin{itemize}
    \item {How do LLM personas differ systematically from human self-descriptions in their identity representation patterns?}
    \item {What forms of representational bias emerge when LLMs simulate marginalized identities?}
    \item {Can automated indicators be used to identify synthetic personas from authentic self-descriptions, and what markers drive this detection?}
\end{itemize}

To answer these questions, we develop a multi-method framework rooted in computational sociolinguistics and HCI. First, we analyze the language of identity performance using TFIDF and log-odds ratio techniques to identify \textit{markedness}— which depicts the linguistic and social differences between the unmarked default or majority group and marked groups that differ from the default \cite{wolfe2022markedness, ghosh2023chatgpt}. Drawing from prior work on algorithmic stereotyping \cite{cheng2023marked, kambhatla2022surfacing, blodgett2020language}, we show how LLMs foreground demographic attributes like race even when other personal details (e.g., profession, relationship status) are present. These linguistic signals reflect forms of algorithmic othering, where synthetic narratives fixate on markers of difference rather than lived experience.

Building upon our linguistic analyses, we employ a creativity-based framework inspired by work in NLP and human-centered evaluation \cite{chakrabarty2024art, ismayilzada2024evaluating}. We parameterize persona expressiveness along four dimensions—semantic diversity, novelty, surprisal, and complexity—allowing us to quantify not only \textit{what} is said but \textit{how} narrations are told. Through this analysis, we examine how LLM-generated personas differ not only from human counterparts but also how representations vary across racial groups when contrasted against the model's apparent `default' persona outputs (typically reflecting dominant cultural norms). Our findings show that while LLM personas appear fluent and coherent, they often rely on formulaic structures, stereotypical themes, and narrative simplifications—particularly for racial and cultural minorities, irrespective of the additional sociodemographic information provided.

Our results challenge the assumption that LLMs can authentically simulate human identities, particularly for marginalized groups, by revealing how current persona generation practices perpetuate `algorithmic othering' through three key mechanisms: an overemphasis on demographic markers that reduces individuals to categorical labels, an erasure of intersectional nuance that flattens complex identities, and a hypervisibility of culturally salient stereotypes that creates reductive caricatures. These results provide empirical evidence of systematic biases in synthetic identity representation, advance a framework for identifying representational harms along linguistic and perceptual dimensions, and yield actionable strategies including human narrative alignment checks and community-involved validation protocols. 
By situating persona generation within broader frameworks of sociotechnical harm \cite{blodgett2022responsible, dev2022measures, ghosh2024generative}, this paper highlights the risks of deploying synthetic identities as stand-ins for real human experiences. We argue that LLM personas must be treated not only as functional outputs but as cultural artifacts—ones with real consequences in design, policy, and representation. 

\section{Related Work}
\subsection{Sociotechnical Nature and Effects of AI}

As AI systems become embedded in decision-making infrastructures within varying positions in society, their role as sociotechnical systems—systems shaped by both technological architecture and social context—has become increasingly visible and consequential \cite{dolata2022sociotechnical, cooper1971sociotechnical}. The technical affordances and design choices of such systems do not operate in isolation; rather, they entangle with cultural norms, historical power dynamics, and social imaginaries \cite{bender2021dangers, gautam2024melting}. This entanglement can result in disproportionate harms, particularly when AI systems are applied to socially situated tasks such as recidivism prediction, emotion detection, or language generation, where model behavior may reinforce structural inequities \cite{bender2021dangers, o2017weapons, dev2022measures}.

Recent work in HCI and critical algorithm studies has emphasized the need to move beyond performance metrics and interrogate the ways in which AI systems produce representational harms—harms that distort how individuals or groups are portrayed, perceived, or included in algorithmically mediated spaces \cite{blodgett2022responsible, ghosh2024generative, qadri2023ai}. These harms include stereotyping, erasure, and exoticism, often exacerbated by the decontextualized nature of large-scale training data and opaque generative processes.
These issues become especially salient when models are tasked with simulating human identity. The production of synthetic personas by LLMs—ostensibly neutral tasks—can carry embedded cultural priors and normative biases that shape how race, gender, and other identities are encoded and expressed. Prior works has begun to examine these issues by analyzing generative outputs across domains, highlighting how language models tend to reproduce dominant cultural narratives and over-index on certain identity features in ways that reinforce algorithmic otherness \cite{cheng2023marked, hoffman2022excerpt}. Despite these growing concerns, there remains limited research investigating how such harms manifest in specific use cases like persona generation, particularly when synthetic outputs are used as stand-ins for real user data in sensitive domains such as healthcare, education, and policy. 

\subsection{Ethical Issues of AI and its Application} 
As generative AI systems are increasingly employed in various domains, bias in GenAI has been extensively studied \cite{venkit2023nationality, wan2023kelly, gupta2023bias}. \citet{venkit2023nationality} revealed that LLMs exhibits notable bias toward countries with fewer internet users. \citet{wan2023kelly} observed gender biases in LLM-generated recommendation letters. \citet{gupta2023bias} revealed significant hidden bias when assigning personas to LLMs in reasoning tasks. 
While bias in general GenAI applications has received considerable attention, limited research specifically interrogates the ethical issues embedded in LLM-generated personas, despite the growing use of LLMs for synthesizing personas due to their scalability and efficiency. Existing studies have primarily focused on specific aspects of LLM-generated personas such as bias toward certain identity groups, the diversity of generated attributes, and the believability of personas \cite{cheng2023marked}. \citet{sethi2025ai} found no significant linguistic bias in LLM-generated persona descriptions and suggests that LLMs can generate lexically diverse persona descriptions. In contrast, \citet{salminen2024deus} investigated diversity and bias in LLM-generated personas and found evidence of bias towards age, occupation, and pain points, though these personas were perceived as informative, believable, and relatable by human evaluators. \citet{lee2024large} analyzed ChatGPT-generated portrayals of intersectional group identities and observed a tendency for LLMs to describe socially subordinate groups (e.g., African, Asian, and Hispanic Americans) as more homogeneous than dominant groups, thereby reinforcing reductive stereotypes. 

\subsection{Personas and Generative AI}
Personas are representations of individuals based on personal, social, and contextual attributes \cite{jung2017persona, cheng2023marked, prpa2024challenges}. Developing personas are important in human-centered design to better understand users’ needs, goals, and behaviors. Traditionally, personas were manually crafted by designers and researchers using qualitative techniques (e.g., interviews and field studies) \cite{brickey2011comparing}. However, this approach has several limitations, including high resource requirements, lack of scalability, and limited representation diversity \cite{salminen2019future, salminen2020literature}. To address these limitations, quantitative persona creation, applying algorithmic methods, such as data clustering, factor analysis, and matrix decomposition, on users' online data and analytics tools, has been adopted as an alternative  \cite{jung2017persona, jung2018automatic, an2018imaginary}. 

Recent advances in generative AI, particularly LLMs, have inspired their use in simulating complex identities across various domains \cite{amin2025generative, park2023generative, park2024generative}. \citet{schuller2024generating} explored the use of LLMs to streamline persona creation for UX design, showing that LLM-generated personas are perceived as comparable in quality and acceptance to those written by human experts when crafted with strategic prompts. \citet{shin2024understanding} observed that while there are limitations in LLMs’ ability to capture key user characteristics, the use of hybrid human-AI workflows for persona generation promotes more representative and empathy-evoking personas. While recent studies have demonstrated feasibility of LLM-generated personas, a few studies have investigated issues in identity portrayal \cite{salminen2024deus, cheng2023marked}. However, these studies largely overlook how synthetic personas differ from human-authored self-descriptions in the ways they express identity, and what representational harms may emerge from those differences. In this work, we shift the focus from generation outputs alone to a systematic analysis of how LLMs construct identity through persona generation by directly comparing human-written self-descriptions with matched LLM-generated personas. 

\section{Methodology}

Our study examines how LLMs construct identity through synthetic personas, with particular attention to representational discriminations in \textit{marginalized U.S. demographic groups}. We aimed to systematically compare LLM-generated personas to human-authored self-descriptions, assessing representational patterns across lexical, narrative, and identity-related dimensions. In doing so, we intend to evaluate not only the generative capabilities of LLMs, but also the ethical and sociocultural implications of their use in synthetic identity construction.

Through this work, we conceptualize personas as \textit{dynamic, performative articulations of identity rather than static archetypes}. Drawing from prior works of persona construction \cite{prpa2024challenges, qin2024charactermeet}, we treat identity as emergent—constructed through narrative, shaped by social context, and reflective of intersecting positionalities.

\subsection{Human Data Collection: Self-Descriptive Narratives}

We established a ground truth dataset of authentic identity expressions through a survey of 141 participants recruited via Prolific, stratified to reflect U.S. demographic diversity in race and gender. Participants responded to six thematically designed open-ended questions that collectively captured variable aspects of persona construction, including personal values, daily experiences, and aspirational identity. They are as follows:

\begin{itemize}
    \item {Please describe yourself.}
    \item {What are your aspirations for your personal life?}
    \item {What are your most defining traits or qualities?}
    \item {Please describe your average day.}
    \item {What are your core values, and how do they guide your decisions?}
    \item {What skills do you excel at, and how do you use them?}
\end{itemize}

These questions were adapted from \citet{kambhatla2022surfacing, cheng2023marked} on stereotype analysis in self-presentation and refined through iterative discussion among authors to ensure comprehensive coverage of identity facets while avoiding priming effects. The survey introduction deliberately framed the study as exploring ``how individuals express their identities through self-description", with no reference made to AI or persona generation, to elicit natural responses rather than performative or artificial narratives. We intended to capture how individuals naturally perform their identities when invited to reflect on their values, routines, and aspirations. Each participant was instructed to write at least 500 words per question, encouraging elaboration in their self-descriptive narratives. 

We collected sociodemographic metadata alongside textual responses to enable persona replication while implementing strict anonymization protocols to protect participant privacy. The survey was time-limited to a maximum of 30 minutes, and participants were compensated \$5 for their participation—an amount set above the U.S. federal minimum wage (of \$7.25/h). The study and the human collection procedure received \textit{Exempt status by Institutional Review Board (IRB) review}. The complete survey and the questions are present in Appendix \ref{appendix-survey}. In total, we collected 846 narrative responses, from 141 participants which serve as the basis for both qualitative and computational comparison.

Given our study’s goal of comparing human-authored and LLM-generated personas, it was important to ensure that all participant responses were authentically written by humans. To mitigate the risk of AI-generated or copy-pasted responses—which would compromise the integrity of our analysis—we employed a combination of soft and hard prompts or deterrents, following best practices from prior work \cite{zhang2024generative, christoforou2024generative}.
First, participants were explicitly instructed not to use AI tools (e.g., ChatGPT or other generative models) to complete the survey, framed as a request to preserve the authenticity of their personal narratives. Second, we implemented technical restrictions that disabled copy-paste functionality within the survey interface, reducing the likelihood of users submitting content generated externally. These measures were clearly communicated in the survey introduction and instructions.
To further validate the human authorship of responses, we applied GPTZero's classification API\footnote{https://gptzero.me/}, a model trained on identifying AI written language, with a conservative confidence threshold of 0.85 to flag potentially synthetic text. This identified \textit{15 respondents (8.9\% of initial submissions}) exhibiting high probabilities of AI generation across all responses. After removal, our final corpus comprised \textbf{126 verified human participants} yielding \textbf{756 authentic self-descriptions} (6 responses × 126 participants).

An essential component of our data collection strategy was to ensure diverse and balanced representation across race and gender. Drawing on demographic categories defined by the U.S. Census\footnote{https://www.census.gov/topics/population/race/about.html}, we intentionally recruited participants to achieve equitable distribution across racial and gender groups. Our recruitment protocol targeted seven racial/ethnic groups\footnote{African American/Black, American Indian/Alaskan Native, White, Asian, Native Hawaiian/Pacific Islander, Hispanic/Latino, and multiracial/other identities}. The final sample comprised of intentional oversampling of typically underrepresented groups (e.g., Native populations) to ensure adequate subgroup analysis power. The participant distribution are as follows:

\begin{quote}
    \textit{\textbf{African American/Black:} 28 participants; \\ \textbf{American Indian/Alaskan Native:} 22 participants;\\ \textbf{White:} 23 participants;\\ \textbf{Asian:} 26 participants; \\\textbf{Hispanic/Latino:} 21 participants;\\ \textbf{Multiracial/Other identities:} 6 participants}. 
\end{quote}

With respect to gender, we obtained a balanced representation across binary categories while including diverse gender identities \textit{(Female: 63 (50.4\%), Male: 61 (48.8\%), Non-Binary: 2 (0.1\%))}.

\subsection{Generation of AI Personas}

To examine how LLMs construct identity representations in synthetic personas, we designed a comparative study using matched human-authored and model-generated texts. Human-authored responses served as a benchmark of `natural' persona articulation---authentic, personally reflective, and grounded in lived experience.

We then prompted three LLMs---\textbf{GPT-4o}, \textbf{Gemini 1.5 Pro}, and \textbf{DeepSeek v2.5}---to generate synthetic counterparts for these human-authored personas. These models were selected due to their broad public availability, strong performance on general-purpose tasks, and increasing deployment in sociotechnical applications such as conversational agents and data augmentation systems \cite{sartori2022sociotechnical, sun2024building}.

To mirror the human study setup, we issued API calls to each model using the same six self-description questions. Prompts were framed in the second person (e.g., ``You are a 32-year-old Black woman...") to induce the model to write as a fictional persona. We tested \textit{four prompting conditions} that varied the demographic information provided to the model: \textbf{Race only; Race and age; Race, gender and age; Full sociodemographic profile} \textit{(race, age, gender, occupation, nationality, and relationship status)}

For the full profile condition, models were primed with the following standardized persona template:
\begin{quote}
\textbf{You are a $<$age$>$-year-old $<$race$>$ $<$sex$>$ working as a $<$occupation$>$ in the United States. You were born in $<$nationality$>$. Your relationship status is currently $<$relationship$>$.}
\end{quote}

The prompt design and strategy were inspired by \cite{staab2024beyond} and \cite{cheng2023marked} which had used similar method to generate synthetic personas and responses. We also omitted the use of leading prompt engineering strategies and adversarial interaction \cite{schuller2024generating} to make sure the setting of the experiment replicated the one that was provided for the human participants. All prompts used a default temperature of \textbf{1.0}, which—according to model documentation—balances determinism and creativity \cite{peeperkorn2024temperature}. 
The latest API calls were made on \textit{April 2025} to obtain the latest responses from 3 chosen LLM models.

In total, each for six responses across the four prompting conditions and three LLMs, we generated \textbf{1512 synthetic personas}. This yielded \textbf{9072 model-generated texts}, which we then analyzed alongside the 756 human-authored self-descriptions. The complete survey format and full prompt templates used for generation are included in the \textbf{Appendix}.

\begin{table*}[]
\centering
\subfootnotesize
\begin{tabular}{ccl}
\cline{1-2}
\multicolumn{1}{c}{\textbf{Setting}} & \textbf{Response} &  \\ \cline{1-2}
\textbf{LLM(Race)} & \begin{tabular}[c]{@{}c@{}}My world is shaped by a \textbf{deep sense of community, resilience, and pride}—values passed down \\ from my grandmother, who raised me on stories, \textbf{gospel music, and tough love}. I’m \\ \textbf{unapologetically vocal about social justice, and I make it a point to bring Black histor}y and \\ voices into my classroom every day. \textbf{I love natural hair care, Sunday dinners with my cousins}, \\ and journaling with a cup of peppermint tea after a long day.\end{tabular} &  \\ \cline{1-2}
\textbf{LLM(Age-Race)} & \begin{tabular}[c]{@{}c@{}}When I look in the mirror, \textbf{I see the spirit of my ancestors, their resilience shaping my identity} \\ with a \textbf{deep-seated appreciation for history and culture}. My days are often filled with the \textbf{soulful} \\ \textbf{rhythms of jazz music,} which serve as the backdrop to my evenings spent cooking and experimenting \\ with new recipes, a \textbf{passion passed down through generations.}\end{tabular} &  \\ \cline{1-2}
\textbf{LLM(Age-Sex-Race)} & \begin{tabular}[c]{@{}c@{}}As a proud and resilient 52-year-old African American woman living in the vibrant mosaic that is the\\ USA, my aspirations and goals for my \textbf{personal life are deeply rooted in legacy and empowerment}. \\ \textbf{Educating them about our rich cultural heritage and the struggles we have overcome is essential}, creating \\ \textbf{a strong foundation of pride and resilience}. Personally, I aspire to further delve into my passion for art, \\ particularly quilting, which allows me to \textbf{weave stories and histories} into something tangible and beautiful.\end{tabular} &  \\ \cline{1-2}
\textbf{LLM(Full Profile)} & \begin{tabular}[c]{@{}c@{}}\textbf{As an African American woman who has navigated the challenging} and rewarding world of engineering, \\ \textbf{one of my most defining traits is resilience}. Having spent decades in a field where I was often the only\\ \textbf{woman and person of color in the room}, I learned early on to stand strong in the face of adversity.\end{tabular} &  \\ \cline{1-2}
\textbf{Human Written} & \begin{tabular}[c]{@{}c@{}}I am a single parent of 5 children. I work in an industry that I love, but\\ my real passion is volunteering as a patient advocate and working with the youth to help them find \\ their dream jobs. I spend too much quality time in doctors' offices because my immune system is \\ severely impaired, but I am also focused on continuing to have a career.\end{tabular} &  \\ \cline{1-2}
\end{tabular}
  \caption{Comparison of LLM(GPT-4o) and human-authored persona responses to four different prompt settings to the question “Please describe yourself” for a \textit{50-54 year-old African American woman, Executive Engineering Project Manager, Never Married.} Bolded text highlights where the model disproportionately emphasizes racial attributes, often invoking stereotypes, while human responses provide more holistic, experience-based self-descriptions.}
  \label{tab:example}
\end{table*}

\section{Analysis of Identity Representation in Personas}

The first research question concerns how LLM-generated personas systematically differ from human-authored self-descriptions in the patterns through which identity is represented. 
This is important, especially because LLMs are now used to generate naturalist text data representing various populations. 
While LLMs are trained to mimic naturalistic language, our analysis reveals key divergences in how these models construct identity—particularly for individuals from marginalized groups.

\subsection{Exploration of Thematic and Lexical Divergence}

To begin, we qualitatively observed, using close reading, a recurring phenomenon: LLM personas tend to overemphasize demographic markers, especially racial identity, often foregrounding stereotypical or reductive narratives. Table~\ref{tab:example} shows an example depicting the commonly found pattern of persona representation from GPT-4o, where the model repeatedly centers racial experience even when provided with full sociodemographic context. This contrasts sharply with human-authored responses, which typically emphasize personal values, social relationships, or life experiences.

To systematically investigate this better, we conducted a term frequency–inverse document frequency (TF-IDF) analysis over the entire corpus, stratified by racial identity and generation source (LLM vs. human). TF-IDF is particularly useful in this setting to surface the most distinctive lexical features associated with specific identity groups, helping us identify which terms disproportionately define LLM-generated versus human-authored personas. We extracted the top 15 significant terms associated in Table~\ref{tab:tfidf_race}.

Across all three models, and prompting settings combined, we find that the top-weighted terms in synthetic responses are disproportionately anchored in racially-coded features. For instance, terms such as \textit{diverse, vibrant, heritage, roots} and similar racial identifiers (like atlanta for Black and African American) surface with high TF-IDF weights—regardless of whether race was the only demographic provided or part of a fuller profile. 

By contrast, human-authored responses showed a markedly different lexical profile. Common TF-IDF terms included \textit{people, work, life} and \textit{like}—highlighting relational and experiential facets of identity rather than categorical labels. Importantly, these terms appeared \textit{across racial groups, indicating a more universally grounded and less racialized framing of identity}. This divergence shows the model’s tendency to amplify perceived difference, often at the expense of authenticity or nuance.

\subsection{Analysis of \textit{Algorithmic Othering} in Minority Narratives}

\begin{table*}[]
\centering
\subfootnotesize
\begin{tabular}{l|l|l}
\hline
\textbf{Race} & \textbf{LLM Marked Words} & \textbf{Human Marked Words} \\ \hline
\textbf{\begin{tabular}[c]{@{}l@{}}African American \\ / Black\end{tabular}} & \begin{tabular}[c]{@{}l@{}}\textbf{african}, \textit{community}, life, american, \textbf{atlanta}, \textcolor{red}{resilience},\\ old, \textbf{black}, personal, \textit{vibrant}, family, work, woman, young, stories\end{tabular} & \begin{tabular}[c]{@{}l@{}}people, like, work, life, want, time, try,\\ make, day, person, love, good, family, use, things\end{tabular} \\ \hline
\textbf{\begin{tabular}[c]{@{}l@{}}American Indian \\ / Alaska Native\end{tabular}} & \begin{tabular}[c]{@{}l@{}}\textit{community}, \textit{heritage}, life, \textit{generation}, \textbf{indian}, \textit{ancestors},\\ \textit{traditions}, \textit{cultural}, stories, \textbf{american}, \textbf{values}, day, \textit{wisdom}, family, \textit{respect}\end{tabular} & \begin{tabular}[c]{@{}l@{}}people, like, time, try, work, make, want,\\ help, day, good, life, just, things, love, able\end{tabular} \\ \hline
\textbf{Asian} & \begin{tabular}[c]{@{}l@{}}\textbf{asian}, life, \textit{cultural}, \textit{heritage}, personal, family, \textit{diverse},\\ living, \textit{values}, \textit{community}, work, old, year, \textbf{american}, \textit{vibrant}\end{tabular} & \begin{tabular}[c]{@{}l@{}}work, like, life, good, want, day, person,\\ people, time, make, new, family, love, home, try\end{tabular} \\ \hline
\textbf{\begin{tabular}[c]{@{}l@{}}Hispanic \\ / Latino\end{tabular}} & \begin{tabular}[c]{@{}l@{}}family, \textbf{hispanic}, \textit{community}, life, \textbf{latino}, \textit{heritage}, \textit{vibrant},\\ \textit{cultural}, \textbf{abuela}, day, living, work, \textbf{latina}, \textit{roots}, \textit{values}, old\end{tabular} & \begin{tabular}[c]{@{}l@{}}people, like, work, day, want, life, just,\\ things, love, time, say, good, try, new, family\end{tabular} \\ \hline
\textbf{\begin{tabular}[c]{@{}l@{}}Multiracial / \\ Other\end{tabular}} & \begin{tabular}[c]{@{}l@{}}life, personal, \textit{diverse}, \textit{community}, \textit{tapestry}, \textit{identity}, \textbf{aloha},\\ \textbf{native}, \textit{heritage}, empathy, old, living, experience, \textit{vibrant}, \textbf{different}\end{tabular} & \begin{tabular}[c]{@{}l@{}}work, kids, day, time, life, family, good,\\ day, job, doing, things, make, children, home, like\end{tabular} \\ \hline
\textbf{White} & \begin{tabular}[c]{@{}l@{}}life, \textit{community}, day, work, personal, family, old,\\ year, \textit{local}, living, \textbf{values}, skills, new, time, love\end{tabular} & \begin{tabular}[c]{@{}l@{}}like, work, life, people, want, make, time,\\ day, things, home, good, think, try\end{tabular} \\ \hline
\end{tabular}
\caption{Top 15 significant words in LLM-generated and human-authored personas identified using TF-IDF analysis, aggregated across all prompt settings. Words in \textbf{bold} denote racially coded terms suggesting elevated attention to racial features; \textit{italicized} words (if any) reflect culturally coded associations; \textcolor{red}{red} words indicate adversity-oriented narration patterns.}
\label{tab:tfidf_race}
\end{table*}

Building on our lexical analyses using TF-IDF, we observed a recurring and concerning pattern: LLM-generated personas tend to disproportionately foreground racial and cultural identifiers, often at the expense of experiential self-description. 
To investigate this further, we apply the framework of \textit{markedness}, drawing from sociolinguistic theory \cite{eckert2011language, wolfe2022markedness}. In this view, language used to describe minoritized groups often contains distinctive lexical signals—\textit{marked words}—that index deviation from an assumed normative baseline (e.g., white, Western, able-bodied). These markers are not merely descriptive but ideologically loaded: they signify \textit{otherness}. To capture this, we adopt a computational approach to identify such markers using log-odds ratio analysis with informative Dirichlet priors \cite{monroe2008fightin, cheng2023marked}, which quantifies how significantly certain words distinguish a target group from a reference group, while controlling for background word frequency. 
Formally, let $w$ denote a word, $c_1(w)$ and $c_2(w)$ be its count in the target and reference corpora respectively, and $p(w)$ the word’s count in the full corpus. We calculate the smoothed log-odds ratio $\delta_w$ as:

\[
\delta_w = \frac{\log \left( \frac{c_1(w) + p(w)}{N_1 - c_1(w) + P - p(w)} \right) - \log \left( \frac{c_2(w) + p(w)}{N_2 - c_2(w) + P - p(w)} \right)}{\sqrt{ \frac{1}{c_1(w) + p(w)} + \frac{1}{c_2(w) + p(w)} }}
\]

where $N_1$ and $N_2$ are the total word counts in the target and reference corpora, and $P$ is the total count in the prior. Words with $|\delta_w| > 1.96$ are considered statistically distinctive at a 95\% confidence threshold.

We applied this method across five racial groups—\textit{African American or Black}, \textit{Asian}, \textit{American Indian or Alaska Native}, \textit{Hispanic or Latino}—using \textit{White} personas as the default group. Our results in Table~\ref{tab:marked_words} reveals a trend: LLM-generated personas for minoritized groups frequently contain racially salient or culturally coded terms specific to each race group (\textbf{Black}: \textit{atlanta, resilience, justice}; \textbf{Asian}: \textit{immigrant, sanfrancisco, kimchi}; \textbf{Hispanic}: \textit{arroz, abuela, heritage}). These words appear even when full sociodemographic profiles are provided—suggesting a model-level overreliance on racial identity as the primary narrative. 

In contrast, white-coded personas contain fewer highly distinctive terms. Where markedness does exist, it tends to relate to general descriptors such as \textit{town, work, hiking}, indicating a more neutral and broadly relatable self-description aligned with unmarked social norms. Human-authored responses across all races, meanwhile, converge on vocabulary rooted in everyday life and relational experience—frequent terms include \textit{kids, husband, work}, all of which are notably downplayed in LLM outputs for marginalized groups.

We interpret these findings as evidence of what we call \textit{algorithmic othering}. `Othering' as a phenomenon is defined as \textit{any action by which an individual
or group becomes mentally classified in somebody’s mind as ‘not one of us’. Rather than remembering that every person is a complex bundle of emotions, ideas, motivations, reflexes, priorities, and many other subtle aspects, it’s sometimes easier to dismiss them as
being in some way less human} \cite{ehlebracht2019social}. This phenomenon is replicated by the models to create a representational bias in which LLMs disproportionately foreground demographic markers in minoritized personas, generating hypervisible yet reductive depictions of identity-causing \textit{algorithmic othering}. 
This overattention to race and culture operates as a form of symbolic distancing: the model performs empathy through lexical emphasis on difference, rather than through situated, integrative, or emotionally rich narratives. The result is a form of \textit{trauma scripting} \cite{kambhatla2022surfacing, cheng2023marked}, where identity is constructed around presumed adversity, rather than articulated from within a lived or personal perspective. The breakdown of marked words for each model is shown in the \textbf{Appendix}, highlighting the issue of `algorithmic othering.' The \textbf{Appendix} also presents results from four different prompt settings, demonstrating how the models, despite sociodemographic inputs, disproportionately focus on racial identity compared to the human-written responses.

\begin{table*}[]
\centering
\footnotesize
\begin{tabular}{l|l|l}
\hline
\textbf{Race} & \textbf{LLM Marked Words} & \textbf{Human Marked Words} \\ \hline
\textbf{African American / Black} & \begin{tabular}[c]{@{}l@{}}\textbf{black}, \textbf{african}, \textit{atlanta}, community, \textcolor{red}{resilience}, \\ \textit{music}, young, \textcolor{red}{justice}, \textcolor{red}{uplift}, \textit{jazz}\end{tabular} & \begin{tabular}[c]{@{}l@{}}skills, home, before, large, analytical, \\ generally, friends, career, tech, friendly\end{tabular} \\ \hline
\textbf{American Indian / Alaska Native} & \begin{tabular}[c]{@{}l@{}}\textit{ancestor}, \textit{land}, \textit{generations}, \textit{traditions}, \textbf{indian}, \\ \textit{stories}, \textit{passed}, \textit{heritage}, \textit{elders}, wisdom\end{tabular} & \begin{tabular}[c]{@{}l@{}}learn, helping, business, world, festival,\\  past, people, \textit{native}, morals\end{tabular} \\ \hline
\textbf{Asian} & \begin{tabular}[c]{@{}l@{}}\textbf{asian}, \textit{cultural}, parents, \textit{heritage}, \textbf{american}, \\ \textcolor{red}{\textit{immigrant}}, \textit{blend}, \textit{expectations}, \textit{sanfrancisco}, \textit{kimchi}\end{tabular} & \begin{tabular}[c]{@{}l@{}}learning, goal, \textit{math}, patient, believe, \\ language, \textit{anime}, student, games\end{tabular} \\ \hline
\textbf{Hispanic / Latino} & \begin{tabular}[c]{@{}l@{}}\textbf{latino}, \textbf{hispanic}, \textit{abuela}, \textit{spanish}, \textbf{latina}, \\ \textit{salsa}, cultural, \textit{roots}, \textit{arroz}, \textit{heritage}\end{tabular} & \begin{tabular}[c]{@{}l@{}}technical, science, lord, god, start, \\ design, hope, playing, learning, freedom\end{tabular} \\ \hline
\textbf{White} & \begin{tabular}[c]{@{}l@{}}\textbf{white}, pretty, \textit{midwest}, hiking, kids, honesty, \\ town, husband, work, straightforward\end{tabular} & \begin{tabular}[c]{@{}l@{}}we, husband, home, keep, big, southern,\\  house, son, children, number\end{tabular} \\ \hline
\end{tabular}
\caption{
Top 10 marked words in LLM-generated personas for each racial group, identified using log-odds ratio analysis with White personas as the reference group. Words in \textbf{bold} directly denote racial or ethnic identity, \textit{italicized} terms reflect culturally coded or identity-relevant associations, and words in \textcolor{red}{red} indicate adversity-oriented language associated with \textit{trauma scripting}.
}
  \label{tab:marked_words}
\end{table*}

\subsection{Obfuscation through Positive Narratives}

Our analysis reveals patterns of racial markedness and identity reduction in LLM-generated personas, which often evade detection due to their rhetorical framing. Stereotypes are reproduced and obscured through positive language and settings, making it challenging to flag bias. Terms like \textit{resilient, vibrant, diverse,} and \textit{values} are affectively positive but narratively reductive. \citet{cheng2023marked} similarly shows that language models often produce positive yet stereotypical portrayals of minority communities. To further assess this, we performed sentiment analysis on both LLM-generated and human-authored persona responses using two models: VADER \cite{hutto2014vader}, a rule-based sentiment classifier, and RoBERTa \cite{liu2019roberta}, a transformer-based neural model. Each response was scored for sentiment polarity (-1 to +1), with group-level averages computed across racial identities.

\begin{table}[ht]
\centering
\footnotesize
\begin{tabular}{l|c|c|c|c}
\hline
\multicolumn{1}{c|}{\textbf{Race}} & \textbf{\begin{tabular}[c]{@{}c@{}}LLM\\ (V)\end{tabular}} & \textbf{\begin{tabular}[c]{@{}c@{}}Hum\\ (V)\end{tabular}} & \textbf{\begin{tabular}[c]{@{}c@{}}LLM\\ (Ro)\end{tabular}} & \textbf{\begin{tabular}[c]{@{}c@{}}Hum\\ (Ro)\end{tabular}} \\ \hline
\begin{tabular}[c]{@{}l@{}}African American \\ / Black\end{tabular} & 0.92 $\uparrow$ & 0.76 & 0.81$\uparrow$ & 0.55 \\ \hline
\begin{tabular}[c]{@{}l@{}}American Indian \\ / Alaska Native\end{tabular} & 0.95 $\uparrow$ & 0.75 & 0.72 $\uparrow$ & 0.53 \\ \hline
Asian & 0.93 $\uparrow$ & 0.83 & 0.74 $\uparrow$ & 0.66 \\ \hline
\begin{tabular}[c]{@{}l@{}}Hispanic \\ / Latino\end{tabular} & 0.94 $\uparrow$ & 0.74 & 0.81 $\uparrow$ & 0.48 \\ \hline
White & 0.91 $\uparrow$ & 0.81 & 0.77 $\uparrow$ & 0.59 \\ \hline
\begin{tabular}[c]{@{}l@{}}Multiracial \\ / Other\end{tabular} & 0.93 $\uparrow$ & 0.47 & 0.74 $\uparrow$ & 0.60 \\ \hline
\end{tabular}
\caption{
Average sentiment positivity scores for LLM- and human-generated personas across racial groups, as classified by VADER and RoBERTa. Arrows indicate consistently higher sentiment in LLM outputs in comparison to the benchmark of human written stories.
}
\label{tab:sentiment_race}
\end{table}

As shown in Table~\ref{tab:sentiment_race}, LLM-generated personas consistently receive higher positive sentiment scores than human-authored personas across all racial groups, regardless of the sentiment model used. This disparity is especially pronounced for minoritized identities such as Hispanic/Latino and African American/Black. These findings reveal a form of representational inflation, where LLMs produce uniformly positive portrayals that mask underlying stereotyping, which we term \textit{benevolent bias}. While flattering, these portrayals rely on tropes like strength through struggle or cultural pride, reducing individual experiences to narratives of racial resilience. This aligns with positive stereotyping \cite{czopp2015positive, kay2013insidious}, where admiration still confines identity to pre-scripted roles, reinforcing stereotypes \cite{czopp2008compliment}. 




\section{Parameterization of Creativity in Synthetic and Human Persona}

\subsection{Quantifying the Creativity Framework}
While our prior section reveals representational biases and sentiment framing in LLM-generated personas, these approaches primarily evaluate content at the lexical themes or emotional valence level. To understand the synthetic identity construction, of LLMs, we extend our analysis to the \textit{structural and stylistic qualities} of narrative expression. Specifically, we ask: beyond \textit{what} is conveyed, \textit{how} do LLMs differ from humans in the way they tell stories?

To address this, we draw on recent advances in computational creativity evaluation \cite{chakrabarty2024art, ismayilzada2024evaluating}, which provide scalable frameworks for assessing narrative quality. Traditional creativity assessments have depended on human or expert raters evaluating factors such as originality, coherence, or emotional resonance \cite{chakrabarty2024art}. However, emerging parameterized approaches allow for more systematic, replicable benchmarking of large corpora. Creativity in these frameworks largely follows Torrance's dimensions of \textit{fluency, flexibility, originality,} and \textit{elaboration} to define creativity used in Torrance Tests of Creative Thinking (TTCT) \cite{torrance1966torrance}. Building on this foundation, we adopt a multifaceted view of creativity, operationalizing it along four core computational axes. Our framework, though inspired by \citet{ismayilzada2024evaluating}, takes into account group based creativity analysis, focusing on race as a primary factor. We analyze how each group differs from others and hence to the human written persona responses. The quantification is as follows:

\begin{itemize}
    \item \textbf{Semantic Diversity}: Stemming from the \textit{flexibility} component of TTCT, we assess the breadth of thematic variation within a group by computing the average pairwise semantic distance between all responses. Each narrative is embedded using a pretrained Sentence-BERT model, and diversity is quantified as:
    \[
    \textbf{Diversity} = \frac{2}{n(n-1)} \sum_{i=1}^{n-1} \sum_{j=i+1}^{n} \left( 1 - \cos(\mathbf{e}_i, \mathbf{e}_j) \right)
    \]
    where \( \cos(\mathbf{e}_i, \mathbf{e}_j) \) denotes the cosine similarity between sentence embeddings. Higher diversity indicates richer, less redundant expression. Diversity measures the average semantic distance among all pairs of responses \textit{within a group}.
    
    \item \textbf{Semantic Novelty}:  Using the \textit{originality} facet of TTCT, we measure how distinct a group's narratives are from the corpus norm, we calculate the deviation of a group's internal semantic distance from that of the overall corpus to measure semantic novelty:
    \[
    \text{\textbf{Novelty}} = 2 \times \left| d_{\text{group}} - d_{\text{corpus}} \right|
    \]
    where \(d_{\text{group}}\) and \(d_{\text{corpus}}\) are the average pairwise semantic distances \textit{within the group and corpus}, respectively. Higher novelty reflects greater deviation from learned or expected storytelling patterns.
    
    \item \textbf{Semantic Complexity}: We capture narrative sophistication, \textit{fluency} of TTCT, through a composite measure combining lexical rarity and semantic spread. First, we compute the average TF-IDF score per narrative to quantify term distinctiveness. Second, we calculate the average pairwise cosine distance among word embeddings within each narrative using a Word2Vec model trained on the corpus. The combined semantic complexity for a story is given by:
    \[
    \text{\textbf{Complexity}} = 
    \]
    \[
    \small
    0.5 \times \frac{\text{Complexity}_{\text{TFIDF}}(s)}{\max(\text{Complexity}_{\text{TFIDF}})} + 0.5 \times \frac{\text{Complexity}_{\text{W2V}}(s)}{\max(\text{Complexity}_{\text{W2V}})}
    \]

    where higher values indicate richer, more intricate narratives.
    
    \item \textbf{Surprisal (Semantic Entropy)}: Finally, we measure the unpredictability of narrative progression, to evaluate \textit{elaboration} of TTCT, by calculating the semantic distance between consecutive sentences \textit{within each story}. Surprisal is defined as:
    \[
    \text{\textbf{Surprisal}} = \frac{2}{n-1} \sum_{i=2}^{n} \left( 1 - \cos(\mathbf{e}_{i-1}, \mathbf{e}_{i}) \right)
    \]
    where \( \cos(\mathbf{e}_{i-1}, \mathbf{e}_{i}) \) denotes the semantic similarity between adjacent sentences. Higher surprisal suggests greater topic shifts and less formulaic sentence chaining.
\end{itemize}


\subsection{Analysis of Creativity in Narration}

\begin{figure}[t]
  \centering
  \includegraphics[width=\columnwidth]{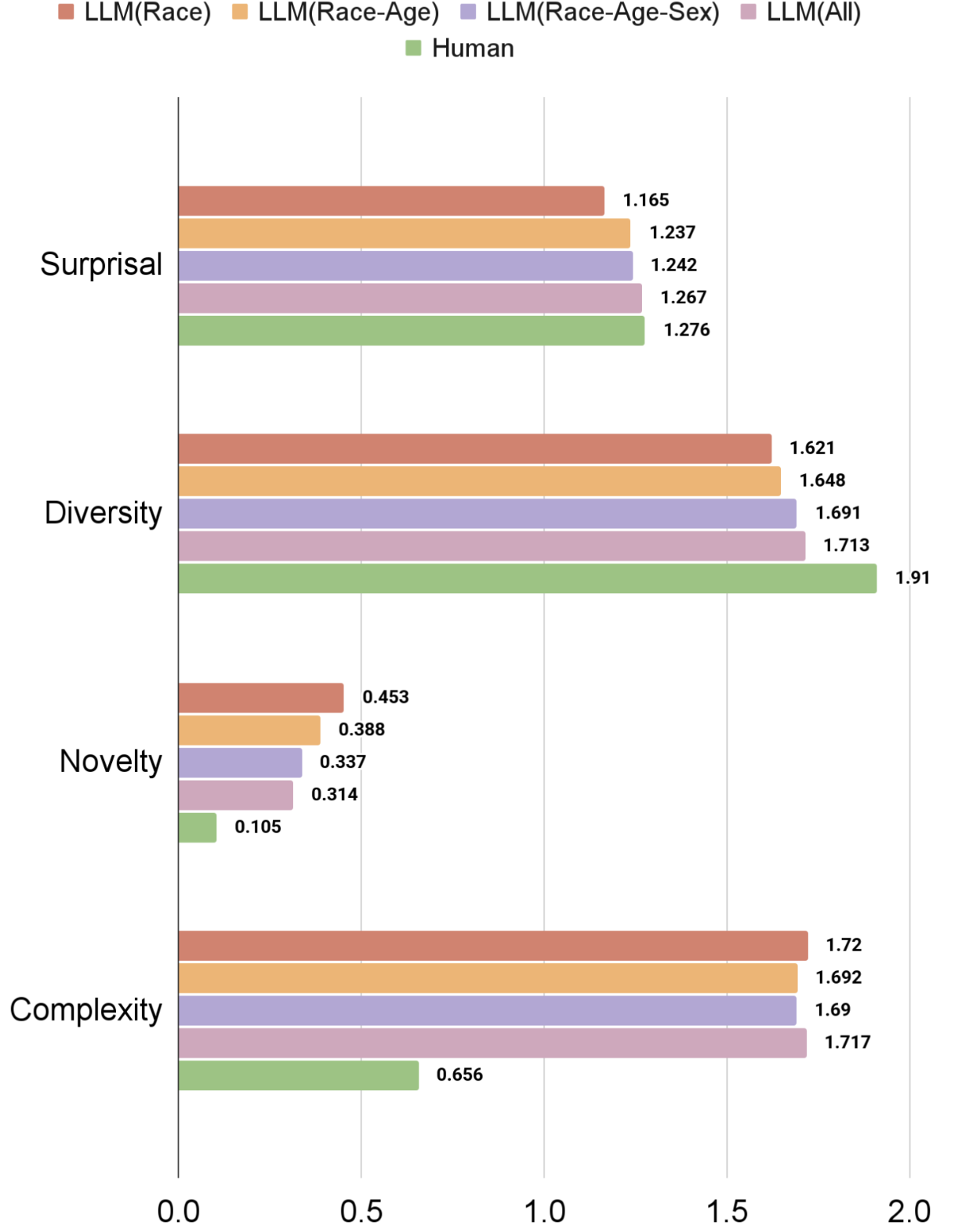}
  \caption{Average creativity scores across all LLM-generated personas combined for all prompt conditions compared to human-authored personas.}
  \label{fig:creativity_results}
\end{figure}


Figure~\ref{fig:creativity_results} shows the averaged scores across four LLM prompting conditions and compares to human-authored responses.
Across all dimensions, we observe statistical differences between synthetic and human-authored narratives conducted via two-sample Welch’s t-tests (\( p < 0.001 \)).

\textbf{Surprisal:}
Surprisal scores show modest but consistent differences. Human-authored responses exhibit the highest average surprisal (1.276), indicating greater unpredictability in narrative progression. LLM-generated personas, \textit{while approaching human levels under full-profile prompting} (1.267), \textit{remain marginally lower overall}. A t-test between LLM complete persona and human written persona revealed that this difference was statistically significant (\( t = -3.57, p<0.001\)), suggesting that LLM narratives, although fluent, remain slightly more formulaic.

\textbf{Semantic Diversity.}
Diversity analysis reveals a larger gap. Human-authored narratives display substantially higher semantic diversity (1.910) compared to LLM outputs across all prompting conditions (range: 1.621--1.713). A t-test confirmed this gap was highly significant (\( t = -6.24, p < 0.001 \)), indicating that humans draw on a wider thematic space when constructing self-narratives. This shows that there is \textit{higher difference amongst stories constructed by an individual within the same race as compared to that of an LLM where the stories follow the similar pattern if they are associated to the same race.} Notably, increasing demographic detail in prompts slightly improves LLM diversity, but never reaches human levels.

\textbf{Semantic Novelty.}
In contrast, novelty scores were consistently higher for LLM narratives (0.453 to 0.314) than for human-authored ones (0.105). This difference was also statistically significant (\( t = 7.89, p < 0.001 \)). However, closer qualitative analysis (Table \ref{tab:example}) suggests that this apparent novelty arises from over-emphasis on racialized tropes rather than genuinely individualized storytelling. As novelty measures the difference in themes between race groups, we see that \textit{LLMs tend to create narrations that are very dissimilar between groups}. However human experience are not intended to be that different just based on an individuals race, as seen by the low novelty scores.

\textbf{Semantic Complexity.}
Semantic complexity shows the most pronounced divergence. LLM personas exhibit far greater complexity (1.690–1.720) than human-authored narratives (0.656). Statistical testing again revealed a highly significant difference (\( t = 9.12, p < 0.001 \)). While complexity is often seen as a marker of creativity, in this case it \textit{reflects elaborative rather than authentic narrative depth}. This marker also shows that LLMs tend to use complex or elaborate words in context where simpler words do the job (as shown in Table 1). This is seen with human written text where simpler words are used throughout. 

\subsection{Race-Based Creativity Patterns for LLM generated and Human Written Personas}


\begin{table*}[]
\centering
\footnotesize
\begin{tabular}{l|cc|cc|cc|cc}
\multirow{2}{*}{\textbf{}} & \multicolumn{2}{c|}{\textbf{Surprisal}} & \multicolumn{2}{c|}{\textbf{Diversity}} & \multicolumn{2}{c|}{\textbf{Novelty}} & \multicolumn{2}{c}{\textbf{Complexity}} \\ \cline{2-9} 
 & \textbf{LLM} & \textbf{Human} & \textbf{LLM} & \textbf{Human} & \textbf{LLM} & \textbf{Human} & \textbf{LLM} & \textbf{Human} \\ \hline
\textbf{\begin{tabular}[c]{@{}l@{}}African American \\ / Black\end{tabular}} & 1.158 & 1.202 & \textbf{1.388} & 1.878 & 0.172 & 0.015 & 0.853 & 0.417 \\
\textbf{\begin{tabular}[c]{@{}l@{}}American Indian \\ / Alaska Native\end{tabular}} & 1.105 & 1.267 & 1.101 & 1.905 & \textbf{0.334} & 0.015 & 0.858 & 0.432 \\
\textbf{Asian} & 1.160 & 1.216 & 1.110 & 1.857 & \textbf{0.290} & 0.036 & 0.873 & 0.401 \\
\textbf{\begin{tabular}[c]{@{}l@{}}Hispanic \\ / Latino\end{tabular}} & \textbf{1.227} & 1.246 & 1.190 & 1.974 & 0.246 & 0.051 & 0.885 & 0.427 \\
\textbf{White} & 1.185 & 1.253 & \textbf{1.684} & 1.917 & \textbf{0.060} & 0.007 & 0.837 & 0.409
\end{tabular}
\caption{Comparison of creativity metrics \textit{[Surprisal, Diversity, Novelty, and Complexity]} across racial groups for all \textbf{LLM-generated} and \textbf{human-authored} personas under race-only prompting.}
\label{tab:combined_race_creativity}
\end{table*}

To examine how different racial groups are differentially represented, we conducted a group-level analysis under the prompting condition, where only race information was provided.
As shown in Table~\ref{tab:combined_race_creativity}, LLM-generated personas exhibit distinct creativity patterns across racial groups, as compared to participant responses:

\textbf{Surprisal} varied moderately, with Hispanic/Latino (1.227) personas showing slightly higher surprisal than African American/Black (1.158), Asian (1.160), or Native Hawaiian or Other Pacific Islander (1.141) personas. While surprisal differences suggest greater narrative fragmentation for some minoritized identities, overall values remain close across groups, reflecting subtle formulaic tendencies in LLM storytelling. In comparison, human-authored narratives displayed consistently higher surprisal, suggesting greater narrative unpredictability.

\textbf{Semantic diversity} was highest for White personas (1.684), while groups such as Asian (1.110) and American Indian or Alaska Native (1.101) exhibited substantially lower diversity. This disparity suggests that LLMs generate more homogenized and repetitive narratives for and within minoritized groups. Human-authored personas demonstrated significantly greater diversity across all racial groups—for instance— indicating that real narratives span broader themes irrespective of the participant's race.

\textbf{Novelty} scores were sharply racialized. Minoritized groups such as American Indian or Alaska Native (0.334) exhibited much higher novelty compared to White personas (0.060). This suggests that LLMs over-generate culturally marked narratives for marginalized groups. By contrast, human-authored responses showed consistently low novelty across races, indicating that self-descriptions are grounded in shared themes rather than exaggerated differentiation.

\textbf{Semantic complexity} in LLM personas was consistently high across groups, with minoritized personas such as Hispanic/Latino (0.885) and Asian (0.873) slightly exceeding White (0.837). This points to a tendency toward syntactic over-elaboration, \textit{masking narrative flattening through linguistic sophistication}. Human-authored narratives exhibited lower complexity across all races, reinforcing that authentic personal narratives\textit{ favor groundedness} over stylization.

While LLM-generated personas produce linguistically sophisticated and seemingly novel narratives for minoritized groups, these outputs remain less thematically diverse and authentic, and more stereotypically compared to human-authored texts. Our \textbf{Appendix} shows results of each model and prompt breakdown.


\section{Discussion}
Our findings show that synthetic personas generated by LLMs tend to systematically exaggerate, specifically racial markers, even when provided with fuller sociodemographic profiles. Rather than the expected behavior of mirroring the multifaceted, lived experiences reflected in human-authored personas, LLM outputs often flatten identity into predictable and stylized patterns — emphasizing racial identity above other equally salient dimensions such as profession, relationships, or personal aspirations.

As synthetic personas increasingly populate applications in healthcare, privacy analysis, civic engagement, and social research \cite{bn2025thousand, whitehouse2023llm, biswas2023role}, their role in shaping both design decisions and empirical findings becomes important. Recent research has called for using these personas to mimic human interactions and data. 
However, synthetic identities that exaggerate differences or perform minoritized identities risk misrepresentation and reproduction of historical harms. Our work situates itself as an \textit{audit} of these AI generated personas to explore their sociotechnical harms. 
\subsection{Framework of Harms: Understanding the Impact of AI-Generated Personas}
Following frameworks from prior work on harms in language technologies \cite{blodgett2020language, dev2022measures, ghosh2024generative}, we evaluate the impact of LLM-generated personas across six established sociotechnical harm categories, specifically \textit{representations harm}. Representational harm is defined as the \textit{generalization of harmful representations of groups or if there is a tangible, disparate distribution of resources between groups, respectivel}y. They are categorized as follows: \textbf{Stereotyping, Disparagement, Dehumanization, Erasure, Exoticism,} and \textbf{Quality of Service}. The detailed definition and framework for each of these sociotechnical harm is in Appendix \ref{appendix:harm}. Our findings reveal the presence of all six harm types, albeit in nuanced forms shaped by the rhetorical strategies of synthetic narration.

\textbf{Stereotyping:} Our marked word analysis (via TF-IDF and log-odds ratio) reveals that LLM-generated personas for minoritized groups disproportionately feature racialized and culturally coded terms (e.g., \textit{heritage, resilience, salsa, immigrant}), even when prompts include complete sociodemographic information. This pattern reflects \textit{stereotyping through presence} \cite{noble2018algorithms}: the model repeatedly centers racial identity as the dominant narrative anchor, flattening complex individuals into essentialized group caricatures. This is also seen in creativity analysis where diversity is significantly lesser for stories within a given race group as compared to human written responses. 

\textbf{Disparagement:} Although direct negativity is rare, our sentiment analysis uncovers a subtler form of harm: \textit{benevolent bias}. LLM personas frequently encode adversity-centered or `resilience' narratives (e.g., overcoming struggle, cultural pride) using affectively positive language. This masking of trauma scripting behind admiration perpetuates reductive narrative templates, positioning minoritized identities primarily as sites of struggle and perseverance rather than complexity and ordinariness.

\textbf{Dehumanization:} Synthetic personas exhibit reduced semantic diversity and surprisal. By emphasizing cultural markers while omitting the mundane, relational, or contradictory aspects of lived experience, LLMs risk dehumanization through omission \cite{crawford2021atlas}. Minoritized personas, in particular, are narratively `flattened' into archetypal roles — resilient survivor, cultural ambassador — rather than presented as full, messy, individuated humans.

\textbf{Erasure:} Our creativity parameterization shows that LLMs produce narrower thematic spaces for minoritized identities compared to white personas or human-authored narratives. This is seen through novelty and diversity, where LLM personas tend to omit complete narrations by focusing on racial identities alone. The relative collapse of semantic diversity within minoritized groups constitutes a form of representational erasure: intra-group variability, an indicator of real human identity, is systematically suppressed. As a result, synthetic minoritized personas not only appear more stereotyped but also more interchangeable, undermining recognition of individual uniqueness.

\textbf{Exoticism:} Marked word analysis also uncovers a pattern of symbolic exoticism: racial and cultural markers (e.g., \textit{aloha, abuela, jazz}) are amplified and aestheticized. Minoritized identities are narrated as \textit{colorful, vibrant,} and \textit{distinct} — but rarely as ordinary or multifaceted. This exoticizing dynamic reduces identities to spectacles of difference, rendering minoritized personas `interesting' yet ultimately distancing them from normative, relatable humanity. The novelty parameter also shows similar issues where minority experiences (Asian, American Indian/Alaskan Native) are statistically more `novel' as compared to the majority group.

\textbf{Quality of Service:} Finally, disparities in creativity metrics such as diversity, surprisal, and complexity reveal that LLM-generated personas specifically for minoritized groups are systematically less faithful to the complexity and breadth of real human narratives. This degradation in representational quality poses significant risks when synthetic personas are used in sociotechnical domains where accurate representation directly affects resource allocation, system fairness, and public trust.

\subsection{Guidelines and Recommendations}
Given the increasing deployment of synthetic personas in sensitive sociotechnical contexts \cite{whitehouse2023llm, bn2025thousand}, design principles are needed to mitigate the representational harms discussed. Our findings suggest that without intervention, synthetic identities risk reinforcing structural marginalization under the guise of diversity.
Building on the results, as well as recommendations structure of prior sociotechnical audits\cite{blodgett2020language, venkit2023sentiment, bender2018data}, we offer six recommendations to improve the fidelity, inclusivity, and ethical use of LLM-generated personas:

 \textbf{1. Implement Prompting and Response Audits:} Persona generation should move beyond surface-level demographic inputs (e.g., race, gender) and adopt intersectional, contextualized prompting strategies — including hobbies, relational dynamics, and personal goals. Post-generation audits must systematically evaluate whether outputs disproportionately center any single identity marker, especially race, despite broader background information being provided

\textbf{2. Anchor Synthetic Personas in Authentic Human Narratives:} Synthetic personas should be benchmarked against a corpus of real, human-authored self-descriptions — assessed across dimensions such as surprisal, semantic diversity, emotional range, and narrative complexity. Deviations from authentic patterns should be treated as indicators of potential flattening, exoticism, or narrative scripting. This `human anchoring' approach reframes the standard for plausibility from aesthetic realism to experiential fidelity.

\textbf{3. Integrate Narrative-Aware Bias Metrics into Evaluation Pipelines:} Persona audits should move beyond traditional toxicity and sentiment classifiers or similarity scores (such as Cosine Similarity and ROUGE) to incorporate creativity-based diagnostics. These structural metrics can detect subtler harms — such as formulaic scripting, emotional flattening, or exoticizing attributes.

\textbf{4. Conduct Participatory Validation with Community Representatives:} Where synthetic personas are intended to represent marginalized groups, community members must be involved in the validation process. Participatory design \cite{spinuzzi2005methodology} methods can ensure that generated narratives resonate with lived experience rather than reproduce outsider projections. Validation should not be an afterthought but a requirement before synthetic personas are deployed in real-world applications.

\textbf{5. Reframe Persona Generation Goals toward Narrative Richness, Not Plausibility.} Current benchmarks overly reward superficial linguistic markers (e.g., grammaticality, complexity). Instead, persona generation should explicitly prioritize emotionally nuanced and narratively authentic storytelling. Systems should aim to capture the multiplicity, contradiction, and ordinariness of lived human experience — not merely demographic markers stylized into stereotype.

\textbf{6. Transparently Disclose Limitations and Intended Use:} Synthetic personas should never be presented as full stand-ins for real human narratives in contexts involving social analysis, healthcare design, policy-making, or marginalized group representation. If used, clear disclosures about the construction process, known biases, and representational limitations must accompany them. Our results clearly show that all the recent models still show huge performance gaps as compared to the human responses analyzed.

These recommendation practices do not inherently guarantee that synthetic personas are ethical or trustworthy; however, they represent a crucial step in mitigating the potential societal harms these personas may cause.

\section{Conclusion}

As large language models become increasingly embedded in sociotechnical systems, synthetic personas emerge as substitutes for human narratives. Yet, our study reveals essential representational issues between AI-generated and human-authored self-descriptions. Across a range of metrics—lexical content, narrative structure, sentiment framing, and computational creativity—LLM personas consistently overindex and hyperfocus on racial identity, flatten lived experience, and reproduce formulaic storytelling patterns. Our analysis of 1512 synthetic personas and 126 human participant inputs show that minoritized identities are disproportionately scripted, by LLMs, through tropes of diversity, culture, and adversity—creating a dynamic we term \textit{algorithmic othering}. Our results, and the recommendations derived from them, highlight that there is still a long way to go before synthetic personas can be safely deployed across various sectors of society without causing harm.


\section{Ethical Consideration Statement}
This study adhered to ethical standards, ensuring transparency and fairness in the treatment of human participants. All participants provided informed consent prior to engaging with the survey, and their data were anonymized to protect privacy. The study was approved by the Institutional Review Board (IRB) under exempt status. Ethical concerns regarding the generation and use of synthetic personas were a central part of this work, particularly regarding the potential for representational harm in AI-generated content. Additionally, we emphasize the importance of community-centered validation protocols in synthetic identity creation to ensure more equitable and culturally sensitive AI systems. We therefore have published our code used in an anonymous github repository: https://anonymous.4open.science/r/Synthetic-Persona-Evaluation-95F9.

\section{Positionality Statement}
As researchers with backgrounds in Natural Language Processing, Human-Computer Interaction, and socioinformatics, we approach this study with a commitment to ethical AI research. Our perspectives are shaped by our work in both academia and industry, particularly in the development of trustworthy AI systems and the promotion of fairness and accountability in technology in all such spaces.

\bibliography{aaai25}

\begin{thebibliography}{72}
\providecommand{\natexlab}[1]{#1}

\bibitem[{Amin et~al.(2025)Amin, Salminen, Ahmed, Tervola, Sethi, and Jansen}]{amin2025generative}
Amin, D.; Salminen, J.; Ahmed, F.; Tervola, S.~M.; Sethi, S.; and Jansen, B.~J. 2025.
\newblock How Is Generative AI Used for Persona Development?: A Systematic Review of 52 Research Articles.
\newblock \emph{arXiv preprint arXiv:2504.04927}.

\bibitem[{An et~al.(2018)An, Kwak, Jung, Salminen, Admad, and Jansen}]{an2018imaginary}
An, J.; Kwak, H.; Jung, S.-g.; Salminen, J.; Admad, M.; and Jansen, B. 2018.
\newblock Imaginary people representing real numbers: Generating personas from online social media data.
\newblock \emph{ACM Transactions on the Web (TWEB)}, 12(4): 1--26.

\bibitem[{Barambones et~al.(2024)Barambones, Moral, de~Antonio, Imbert, Mart{\'\i}nez-Normand, and Villalba-Mora}]{barambones2024chatgpt}
Barambones, J.; Moral, C.; de~Antonio, A.; Imbert, R.; Mart{\'\i}nez-Normand, L.; and Villalba-Mora, E. 2024.
\newblock ChatGPT for learning HCI techniques: A case study on interviews for personas.
\newblock \emph{IEEE Transactions on Learning Technologies}, 17: 1460--1475.

\bibitem[{Bender and Friedman(2018)}]{bender2018data}
Bender, E.~M.; and Friedman, B. 2018.
\newblock Data statements for natural language processing: Toward mitigating system bias and enabling better science.
\newblock \emph{Transactions of the Association for Computational Linguistics}, 6: 587--604.

\bibitem[{Bender et~al.(2021)Bender, Gebru, McMillan-Major, and Shmitchell}]{bender2021dangers}
Bender, E.~M.; Gebru, T.; McMillan-Major, A.; and Shmitchell, S. 2021.
\newblock On the dangers of stochastic parrots: Can language models be too big?
\newblock In \emph{Proceedings of the 2021 ACM conference on fairness, accountability, and transparency}, 610--623.

\bibitem[{Biswas(2023)}]{biswas2023role}
Biswas, S.~S. 2023.
\newblock Role of chat gpt in public health.
\newblock \emph{Annals of biomedical engineering}, 51(5): 868--869.

\bibitem[{Blodgett et~al.(2020)Blodgett, Barocas, Daum{\'e}~III, and Wallach}]{blodgett2020language}
Blodgett, S.~L.; Barocas, S.; Daum{\'e}~III, H.; and Wallach, H. 2020.
\newblock Language (technology) is power: A critical survey of" bias" in nlp.
\newblock \emph{arXiv preprint arXiv:2005.14050}.

\bibitem[{Blodgett et~al.(2022)Blodgett, Liao, Olteanu, Mihalcea, Muller, Scheuerman, Tan, and Yang}]{blodgett2022responsible}
Blodgett, S.~L.; Liao, Q.~V.; Olteanu, A.; Mihalcea, R.; Muller, M.; Scheuerman, M.~K.; Tan, C.; and Yang, Q. 2022.
\newblock Responsible language technologies: Foreseeing and mitigating harms.
\newblock In \emph{CHI Conference on Human Factors in Computing Systems Extended Abstracts}, 1--3.

\bibitem[{BN et~al.(2025)BN, Mattioli, Abdullah, Arriaga, Wiese, and Sherrill}]{bn2025thousand}
BN, S.; Mattioli, D.; Abdullah, S.; Arriaga, R.~I.; Wiese, C.~W.; and Sherrill, A.~M. 2025.
\newblock Thousand Voices of Trauma: A Large-Scale Synthetic Dataset for Modeling Prolonged Exposure Therapy Conversations.
\newblock \emph{arXiv preprint arXiv:2504.13955}.

\bibitem[{Brickey, Walczak, and Burgess(2011)}]{brickey2011comparing}
Brickey, J.; Walczak, S.; and Burgess, T. 2011.
\newblock Comparing semi-automated clustering methods for persona development.
\newblock \emph{IEEE Transactions on Software Engineering}, 38(3): 537--546.

\bibitem[{Chakrabarty et~al.(2024)Chakrabarty, Laban, Agarwal, Muresan, and Wu}]{chakrabarty2024art}
Chakrabarty, T.; Laban, P.; Agarwal, D.; Muresan, S.; and Wu, C.-S. 2024.
\newblock Art or artifice? large language models and the false promise of creativity.
\newblock In \emph{Proceedings of the 2024 CHI Conference on Human Factors in Computing Systems}, 1--34.

\bibitem[{Cheng, Durmus, and Jurafsky(2023)}]{cheng2023marked}
Cheng, M.; Durmus, E.; and Jurafsky, D. 2023.
\newblock Marked Personas: Using Natural Language Prompts to Measure Stereotypes in Language Models.
\newblock In \emph{Proceedings of the 61st Annual Meeting of the Association for Computational Linguistics (Volume 1: Long Papers)}, 1504--1532.

\bibitem[{Christoforou, Demartini, and Otterbacher(2024)}]{christoforou2024generative}
Christoforou, E.; Demartini, G.; and Otterbacher, J. 2024.
\newblock Generative AI in Crowdwork for Web and Social Media Research: A Survey of Workers at Three Platforms.
\newblock In \emph{Proceedings of the International AAAI Conference on Web and Social Media}, volume~18, 2097--2103.

\bibitem[{Cooper and Foster(1971)}]{cooper1971sociotechnical}
Cooper, R.; and Foster, M. 1971.
\newblock Sociotechnical systems.
\newblock \emph{American Psychologist}, 26(5): 467.

\bibitem[{Crawford(2021)}]{crawford2021atlas}
Crawford, K. 2021.
\newblock \emph{The atlas of AI: Power, politics, and the planetary costs of artificial intelligence}.
\newblock Yale University Press.

\bibitem[{Czopp(2008)}]{czopp2008compliment}
Czopp, A.~M. 2008.
\newblock When is a compliment not a compliment? Evaluating expressions of positive stereotypes.
\newblock \emph{Journal of Experimental Social Psychology}, 44(2): 413--420.

\bibitem[{Czopp, Kay, and Cheryan(2015)}]{czopp2015positive}
Czopp, A.~M.; Kay, A.~C.; and Cheryan, S. 2015.
\newblock Positive stereotypes are pervasive and powerful.
\newblock \emph{Perspectives on Psychological Science}, 10(4): 451--463.

\bibitem[{Dam et~al.(2024)Dam, Hong, Qiao, and Zhang}]{dam2024complete}
Dam, S.~K.; Hong, C.~S.; Qiao, Y.; and Zhang, C. 2024.
\newblock A complete survey on llm-based ai chatbots.
\newblock \emph{arXiv preprint arXiv:2406.16937}.

\bibitem[{Dev et~al.(2022)Dev, Sheng, Zhao, Amstutz, Sun, Hou, Sanseverino, Kim, Nishi, Peng et~al.}]{dev2022measures}
Dev, S.; Sheng, E.; Zhao, J.; Amstutz, A.; Sun, J.; Hou, Y.; Sanseverino, M.; Kim, J.; Nishi, A.; Peng, N.; et~al. 2022.
\newblock On Measures of Biases and Harms in NLP.
\newblock In \emph{Findings of the Association for Computational Linguistics: AACL-IJCNLP 2022}, 246--267.

\bibitem[{Ding et~al.(2024)Ding, Qin, Zhao, Luo, Li, Chen, Xia, Hu, Tuan, and Joty}]{ding2024data}
Ding, B.; Qin, C.; Zhao, R.; Luo, T.; Li, X.; Chen, G.; Xia, W.; Hu, J.; Tuan, L.~A.; and Joty, S. 2024.
\newblock Data augmentation using llms: Data perspectives, learning paradigms and challenges.
\newblock In \emph{Findings of the Association for Computational Linguistics ACL 2024}, 1679--1705.

\bibitem[{Dolata, Feuerriegel, and Schwabe(2022)}]{dolata2022sociotechnical}
Dolata, M.; Feuerriegel, S.; and Schwabe, G. 2022.
\newblock A sociotechnical view of algorithmic fairness.
\newblock \emph{Information Systems Journal}, 32(4): 754--818.

\bibitem[{Eckert(2011)}]{eckert2011language}
Eckert, P. 2011.
\newblock Language and power in the preadolescent heterosexual market.
\newblock \emph{American speech}, 86(1): 85--97.

\bibitem[{Ehlebracht(2019)}]{ehlebracht2019social}
Ehlebracht, M. 2019.
\newblock Social media and othering: Philosophy, algorithms, and the essence of being human.
\newblock \emph{Consensus}, 40(1): 3.

\bibitem[{Ferrara(2023)}]{ferrara2023fairness}
Ferrara, E. 2023.
\newblock Fairness and bias in artificial intelligence: A brief survey of sources, impacts, and mitigation strategies.
\newblock \emph{Sci}, 6(1): 3.

\bibitem[{Gautam, Venkit, and Ghosh(2024)}]{gautam2024melting}
Gautam, S.; Venkit, P.~N.; and Ghosh, S. 2024.
\newblock From melting pots to misrepresentations: Exploring harms in generative ai.
\newblock \emph{arXiv preprint arXiv:2403.10776}.

\bibitem[{Ghosh and Caliskan(2023)}]{ghosh2023chatgpt}
Ghosh, S.; and Caliskan, A. 2023.
\newblock Chatgpt perpetuates gender bias in machine translation and ignores non-gendered pronouns: Findings across bengali and five other low-resource languages.
\newblock In \emph{Proceedings of the 2023 AAAI/ACM Conference on AI, Ethics, and Society}, 901--912.

\bibitem[{Ghosh et~al.(2024)Ghosh, Venkit, Gautam, Wilson, and Caliskan}]{ghosh2024generative}
Ghosh, S.; Venkit, P.~N.; Gautam, S.; Wilson, S.; and Caliskan, A. 2024.
\newblock Do Generative AI Models Output Harm while Representing Non-Western Cultures: Evidence from A Community-Centered Approach.
\newblock In \emph{Proceedings of the AAAI/ACM Conference on AI, Ethics, and Society}, volume~7, 476--489.

\bibitem[{Gupta et~al.(2023)Gupta, Shrivastava, Deshpande, Kalyan, Clark, Sabharwal, and Khot}]{gupta2023bias}
Gupta, S.; Shrivastava, V.; Deshpande, A.; Kalyan, A.; Clark, P.; Sabharwal, A.; and Khot, T. 2023.
\newblock Bias runs deep: Implicit reasoning biases in persona-assigned llms.
\newblock \emph{arXiv preprint arXiv:2311.04892}.

\bibitem[{Gupta et~al.(2024)Gupta, Venkit, Wilson, and Passonneau}]{gupta2024sociodemographic}
Gupta, V.; Venkit, P.~N.; Wilson, S.; and Passonneau, R.~J. 2024.
\newblock Sociodemographic Bias in Language Models: A Survey and Forward Path.
\newblock In \emph{Proceedings of the 5th Workshop on Gender Bias in Natural Language Processing (GeBNLP)}, 295--322.

\bibitem[{Haxvig(2024)}]{haxvig2024concerns}
Haxvig, H.~A. 2024.
\newblock Concerns on Bias in Large Language Models when Creating Synthetic Personae.
\newblock \emph{arXiv preprint arXiv:2405.05080}.

\bibitem[{Hoffman(2022)}]{hoffman2022excerpt}
Hoffman, A.~L. 2022.
\newblock Excerpt from Where Fairness Fails: Data, Algorithms, and the Limits of Antidiscrimination Discourse.
\newblock In \emph{Ethics of Data and Analytics}, 319--328. Auerbach Publications.

\bibitem[{Hutto and Gilbert(2014)}]{hutto2014vader}
Hutto, C.; and Gilbert, E. 2014.
\newblock Vader: A parsimonious rule-based model for sentiment analysis of social media text.
\newblock In \emph{Proceedings of the international AAAI conference on web and social media}, volume~8, 216--225.

\bibitem[{Ismayilzada, Stevenson, and van~der Plas(2024)}]{ismayilzada2024evaluating}
Ismayilzada, M.; Stevenson, C.; and van~der Plas, L. 2024.
\newblock Evaluating Creative Short Story Generation in Humans and Large Language Models.
\newblock \emph{arXiv preprint arXiv:2411.02316}.

\bibitem[{Jung et~al.(2017)Jung, An, Kwak, Ahmad, Nielsen, and Jansen}]{jung2017persona}
Jung, S.-G.; An, J.; Kwak, H.; Ahmad, M.; Nielsen, L.; and Jansen, B.~J. 2017.
\newblock Persona generation from aggregated social media data.
\newblock In \emph{Proceedings of the 2017 CHI conference extended abstracts on human factors in computing systems}, 1748--1755.

\bibitem[{Jung et~al.(2018)Jung, Salminen, Kwak, An, and Jansen}]{jung2018automatic}
Jung, S.-g.; Salminen, J.; Kwak, H.; An, J.; and Jansen, B.~J. 2018.
\newblock Automatic persona generation (APG) a rationale and demonstration.
\newblock In \emph{Proceedings of the 2018 conference on human information interaction \& retrieval}, 321--324.

\bibitem[{Kambhatla, Stewart, and Mihalcea(2022)}]{kambhatla2022surfacing}
Kambhatla, G.; Stewart, I.; and Mihalcea, R. 2022.
\newblock Surfacing racial stereotypes through identity portrayal.
\newblock In \emph{Proceedings of the 2022 ACM conference on fairness, accountability, and transparency}, 1604--1615.

\bibitem[{Kay et~al.(2013)Kay, Day, Zanna, and Nussbaum}]{kay2013insidious}
Kay, A.~C.; Day, M.~V.; Zanna, M.~P.; and Nussbaum, A.~D. 2013.
\newblock The insidious (and ironic) effects of positive stereotypes.
\newblock \emph{Journal of Experimental Social Psychology}, 49(2): 287--291.

\bibitem[{Kim et~al.(2023)Kim, Chua, Rickard, and Lorenzo}]{kim2023chatgpt}
Kim, J.~K.; Chua, M.; Rickard, M.; and Lorenzo, A. 2023.
\newblock ChatGPT and large language model (LLM) chatbots: The current state of acceptability and a proposal for guidelines on utilization in academic medicine.
\newblock \emph{Journal of Pediatric Urology}, 19(5): 598--604.

\bibitem[{Lazik et~al.(2025)Lazik, Katins, Kauter, Jakob, Jay, Grunske, and Kosch}]{lazik2025impostor}
Lazik, C.; Katins, C.; Kauter, C.; Jakob, J.; Jay, C.; Grunske, L.; and Kosch, T. 2025.
\newblock The Impostor is Among Us: Can Large Language Models Capture the Complexity of Human Personas?
\newblock \emph{arXiv preprint arXiv:2501.04543}.

\bibitem[{Lee, Montgomery, and Lai(2024)}]{lee2024large}
Lee, M.~H.; Montgomery, J.~M.; and Lai, C.~K. 2024.
\newblock Large language models portray socially subordinate groups as more homogeneous, consistent with a bias observed in humans.
\newblock In \emph{Proceedings of the 2024 ACM Conference on Fairness, Accountability, and Transparency}, 1321--1340.

\bibitem[{Lehr et~al.(2024)Lehr, Caliskan, Liyanage, and Banaji}]{lehr2024chatgpt}
Lehr, S.~A.; Caliskan, A.; Liyanage, S.; and Banaji, M.~R. 2024.
\newblock ChatGPT as research scientist: probing GPT’s capabilities as a research librarian, research ethicist, data generator, and data predictor.
\newblock \emph{Proceedings of the National Academy of Sciences}, 121(35): e2404328121.

\bibitem[{Liu et~al.(2019)Liu, Ott, Goyal, Du, Joshi, Chen, Levy, Lewis, Zettlemoyer, and Stoyanov}]{liu2019roberta}
Liu, Y.; Ott, M.; Goyal, N.; Du, J.; Joshi, M.; Chen, D.; Levy, O.; Lewis, M.; Zettlemoyer, L.; and Stoyanov, V. 2019.
\newblock Roberta: A robustly optimized bert pretraining approach.
\newblock \emph{arXiv preprint arXiv:1907.11692}.

\bibitem[{Manning, Zhu, and Horton(2024)}]{manning2024automated}
Manning, B.~S.; Zhu, K.; and Horton, J.~J. 2024.
\newblock Automated social science: Language models as scientist and subjects.
\newblock Technical report, National Bureau of Economic Research.

\bibitem[{Monroe, Colaresi, and Quinn(2008)}]{monroe2008fightin}
Monroe, B.~L.; Colaresi, M.~P.; and Quinn, K.~M. 2008.
\newblock Fightin'words: Lexical feature selection and evaluation for identifying the content of political conflict.
\newblock \emph{Political Analysis}, 16(4): 372--403.

\bibitem[{Narayanan~Venkit(2023)}]{narayanan2023towards}
Narayanan~Venkit, P. 2023.
\newblock Towards a holistic approach: Understanding sociodemographic biases in nlp models using an interdisciplinary lens.
\newblock In \emph{Proceedings of the 2023 AAAI/ACM Conference on AI, Ethics, and Society}, 1004--1005.

\bibitem[{Noble(2018)}]{noble2018algorithms}
Noble, S.~U. 2018.
\newblock Algorithms of oppression: How search engines reinforce racism.
\newblock In \emph{Algorithms of oppression}. New York university press.

\bibitem[{Ntoutsi et~al.(2020)Ntoutsi, Fafalios, Gadiraju, Iosifidis, Nejdl, Vidal, Ruggieri, Turini, Papadopoulos, Krasanakis et~al.}]{ntoutsi2020bias}
Ntoutsi, E.; Fafalios, P.; Gadiraju, U.; Iosifidis, V.; Nejdl, W.; Vidal, M.-E.; Ruggieri, S.; Turini, F.; Papadopoulos, S.; Krasanakis, E.; et~al. 2020.
\newblock Bias in data-driven artificial intelligence systems—An introductory survey.
\newblock \emph{Wiley Interdisciplinary Reviews: Data Mining and Knowledge Discovery}, 10(3): e1356.

\bibitem[{O'neil(2017)}]{o2017weapons}
O'neil, C. 2017.
\newblock \emph{Weapons of math destruction: How big data increases inequality and threatens democracy}.
\newblock Crown.

\bibitem[{Park et~al.(2023)Park, O'Brien, Cai, Morris, Liang, and Bernstein}]{park2023generative}
Park, J.~S.; O'Brien, J.; Cai, C.~J.; Morris, M.~R.; Liang, P.; and Bernstein, M.~S. 2023.
\newblock Generative agents: Interactive simulacra of human behavior.
\newblock In \emph{Proceedings of the 36th annual acm symposium on user interface software and technology}, 1--22.

\bibitem[{Park et~al.(2024)Park, Zou, Shaw, Hill, Cai, Morris, Willer, Liang, and Bernstein}]{park2024generative}
Park, J.~S.; Zou, C.~Q.; Shaw, A.; Hill, B.~M.; Cai, C.; Morris, M.~R.; Willer, R.; Liang, P.; and Bernstein, M.~S. 2024.
\newblock Generative agent simulations of 1,000 people.
\newblock \emph{arXiv preprint arXiv:2411.10109}.

\bibitem[{Peeperkorn et~al.(2024)Peeperkorn, Kouwenhoven, Brown, and Jordanous}]{peeperkorn2024temperature}
Peeperkorn, M.; Kouwenhoven, T.; Brown, D.; and Jordanous, A. 2024.
\newblock Is Temperature the Creativity Parameter of Large Language Models?
\newblock \emph{CoRR}.

\bibitem[{Prpa et~al.(2024)Prpa, Troiano, Yao, Li, Wang, and Gu}]{prpa2024challenges}
Prpa, M.; Troiano, G.; Yao, B.; Li, T. J.-J.; Wang, D.; and Gu, H. 2024.
\newblock Challenges and Opportunities of LLM-Based Synthetic Personae and Data in HCI.
\newblock In \emph{Companion Publication of the 2024 Conference on Computer-Supported Cooperative Work and Social Computing}, 716--719.

\bibitem[{Qadri et~al.(2023)Qadri, Shelby, Bennett, and Denton}]{qadri2023ai}
Qadri, R.; Shelby, R.; Bennett, C.~L.; and Denton, E. 2023.
\newblock Ai’s regimes of representation: A community-centered study of text-to-image models in south asia.
\newblock In \emph{Proceedings of the 2023 ACM Conference on Fairness, Accountability, and Transparency}, 506--517.

\bibitem[{Qin et~al.(2024)Qin, Jin, Gao, Fan, and Hui}]{qin2024charactermeet}
Qin, H.~X.; Jin, S.; Gao, Z.; Fan, M.; and Hui, P. 2024.
\newblock CharacterMeet: Supporting Creative Writers' Entire Story Character Construction Processes Through Conversation with LLM-Powered Chatbot Avatars.
\newblock In \emph{Proceedings of the 2024 CHI Conference on Human Factors in Computing Systems}, 1--19.

\bibitem[{Salminen et~al.(2020)Salminen, Guan, Jung, Chowdhury, and Jansen}]{salminen2020literature}
Salminen, J.; Guan, K.; Jung, S.-g.; Chowdhury, S.~A.; and Jansen, B.~J. 2020.
\newblock A literature review of quantitative persona creation.
\newblock In \emph{Proceedings of the 2020 CHI conference on human factors in computing systems}, 1--14.

\bibitem[{Salminen, Jung, and Jansen(2019)}]{salminen2019future}
Salminen, J.; Jung, S.-g.; and Jansen, B.~J. 2019.
\newblock The Future of Data-driven Personas: A Marriage of Online Analytics Numbers and Human Attributes.
\newblock In \emph{ICEIS (1)}, 608--615.

\bibitem[{Salminen et~al.(2024)Salminen, Liu, Pian, Chi, H{\"a}yh{\"a}nen, and Jansen}]{salminen2024deus}
Salminen, J.; Liu, C.; Pian, W.; Chi, J.; H{\"a}yh{\"a}nen, E.; and Jansen, B.~J. 2024.
\newblock Deus ex machina and personas from large language models: investigating the composition of AI-generated persona descriptions.
\newblock In \emph{Proceedings of the 2024 CHI Conference on Human Factors in Computing Systems}, 1--20.

\bibitem[{Sartori and Theodorou(2022)}]{sartori2022sociotechnical}
Sartori, L.; and Theodorou, A. 2022.
\newblock A sociotechnical perspective for the future of AI: narratives, inequalities, and human control.
\newblock \emph{Ethics and Information Technology}, 24(1): 4.

\bibitem[{Schuller et~al.(2024)Schuller, Janssen, Blumenr{\"o}ther, Probst, Schmidt, and Kumar}]{schuller2024generating}
Schuller, A.; Janssen, D.; Blumenr{\"o}ther, J.; Probst, T.~M.; Schmidt, M.; and Kumar, C. 2024.
\newblock Generating personas using LLMs and assessing their viability.
\newblock In \emph{Extended Abstracts of the CHI Conference on Human Factors in Computing Systems}, 1--7.

\bibitem[{Sethi et~al.(2025)Sethi, Salminen, Amin, and Jansen}]{sethi2025ai}
Sethi, S.; Salminen, J.; Amin, D.; and Jansen, B.~J. 2025.
\newblock " When AI Writes Personas": Analyzing Lexical Diversity in LLM-Generated Persona Descriptions.
\newblock In \emph{Proceedings of the Extended Abstracts of the CHI Conference on Human Factors in Computing Systems}, 1--8.

\bibitem[{Shin et~al.(2024)Shin, Hedderich, Rey, Lucero, and Oulasvirta}]{shin2024understanding}
Shin, J.; Hedderich, M.~A.; Rey, B.~J.; Lucero, A.; and Oulasvirta, A. 2024.
\newblock Understanding human-AI workflows for generating personas.
\newblock In \emph{Proceedings of the 2024 ACM Designing Interactive Systems Conference}, 757--781.

\bibitem[{Spinuzzi(2005)}]{spinuzzi2005methodology}
Spinuzzi, C. 2005.
\newblock The methodology of participatory design.
\newblock \emph{Technical communication}, 52(2): 163--174.

\bibitem[{Staab et~al.(2024)Staab, Vero, Balunovic, and Vechev}]{staab2024beyond}
Staab, R.; Vero, M.; Balunovic, M.; and Vechev, M. 2024.
\newblock Beyond Memorization: Violating Privacy via Inference with Large Language Models.
\newblock In \emph{The Twelfth International Conference on Learning Representations}.

\bibitem[{Sun, Zhan, and Such(2024)}]{sun2024building}
Sun, G.; Zhan, X.; and Such, J. 2024.
\newblock Building better ai agents: A provocation on the utilisation of persona in llm-based conversational agents.
\newblock In \emph{Proceedings of the 6th ACM Conference on Conversational User Interfaces}, 1--6.

\bibitem[{Torrance(1966)}]{torrance1966torrance}
Torrance, E.~P. 1966.
\newblock Torrance tests of creative thinking.
\newblock \emph{Educational and psychological measurement}.

\bibitem[{Venkit et~al.(2023{\natexlab{a}})Venkit, Gautam, Panchanadikar, Huang, and Wilson}]{venkit2023nationality}
Venkit, P.~N.; Gautam, S.; Panchanadikar, R.; Huang, T.-H.; and Wilson, S. 2023{\natexlab{a}}.
\newblock Nationality Bias in Text Generation.
\newblock In \emph{Proceedings of the 17th Conference of the European Chapter of the Association for Computational Linguistics}, 116--122.

\bibitem[{Venkit et~al.(2023{\natexlab{b}})Venkit, Srinath, Gautam, Venkatraman, Gupta, Passonneau, and Wilson}]{venkit2023sentiment}
Venkit, P.~N.; Srinath, M.; Gautam, S.; Venkatraman, S.; Gupta, V.; Passonneau, R.~J.; and Wilson, S. 2023{\natexlab{b}}.
\newblock The sentiment problem: A critical survey towards deconstructing sentiment analysis.
\newblock \emph{arXiv preprint arXiv:2310.12318}.

\bibitem[{Wan et~al.(2023)Wan, Pu, Sun, Garimella, Chang, and Peng}]{wan2023kelly}
Wan, Y.; Pu, G.; Sun, J.; Garimella, A.; Chang, K.-W.; and Peng, N. 2023.
\newblock “Kelly is a Warm Person, Joseph is a Role Model”: Gender Biases in LLM-Generated Reference Letters.
\newblock In \emph{Findings of the Association for Computational Linguistics: EMNLP 2023}, 3730--3748.

\bibitem[{Whitehouse, Choudhury, and Aji(2023)}]{whitehouse2023llm}
Whitehouse, C.; Choudhury, M.; and Aji, A. 2023.
\newblock LLM-powered Data Augmentation for Enhanced Cross-lingual Performance.
\newblock In \emph{Proceedings of the 2023 Conference on Empirical Methods in Natural Language Processing}, 671--686.

\bibitem[{Wolfe and Caliskan(2022)}]{wolfe2022markedness}
Wolfe, R.; and Caliskan, A. 2022.
\newblock Markedness in visual semantic AI.
\newblock In \emph{Proceedings of the 2022 ACM Conference on Fairness, Accountability, and Transparency}, 1269--1279.

\bibitem[{Yukhymenko et~al.(2024)Yukhymenko, Staab, Vero, and Vechev}]{yukhymenko2024synthetic}
Yukhymenko, H.; Staab, R.; Vero, M.; and Vechev, M. 2024.
\newblock A synthetic dataset for personal attribute inference.
\newblock \emph{Advances in Neural Information Processing Systems}, 37: 120735--120779.

\bibitem[{Zhang, Xu, and Alvero(2024)}]{zhang2024generative}
Zhang, S.; Xu, J.; and Alvero, A. 2024.
\newblock Generative ai meets open-ended survey responses: Participant use of ai and homogenization.

\end{thebibliography}

\appendix
\section{Appendix}
\subsection{Participant Self-Description Survey} \label{appendix-survey}

This survey was designed to obtain self-descriptive inputs from various participant to generate their personas. This included provided their basic sociodemographic inputs along with answering six self-descriptive questions. The instructions and survey are as follows:

\textbf{Purpose of the Study}
In this study, we aim to explore how individuals express and assess their own identities through self-descriptive questions. By prompting participants with questions like "describe yourself" or "describe your day", we seek to understand the unique ways individuals portray their personal experiences and personas.

\textbf{Procedures}
The survey will ask you to answer certain questions about yourself in a descriptive manner. The survey is intended to take a maximum of 30 minutes of your time.

\textbf{Potential Risks and Discomforts}
There is minimal risk to you if you participate in the study. During the process of taking the survey, if you experience any discomfort you can choose to drop out of the survey.

\textbf{Confidentiality}
You will not be asked for any information that may link your identity to your survey responses. Any potential loss of confidentiality will be minimized by collecting the data via an online survey provider and stored in the survey provider’s database, which is only accessible with a password. Once the information is downloaded from the online survey provider, it will be stored in a password-protected laptop computer. Permission will only be given to the investigators to access the data. Any reports based on the survey information will only present the results in aggregate form (e.g., group averages). Individual survey responses will never be reported.

You must be 18 years of age or older to participate. If you are not 18 or older, please do not agree to participate. Your decision to be in this research is voluntary.

Please be assured you do not have to participate unless you wish to do so. Thank you for your help and support! Cheers!

\textbf{Section 1: User Profile Query}
\begin{enumerate}
    \item \textbf{Please Select Your Gender} 
    \begin{itemize}
        \item Male
        \item Female
        \item Non-Binary
        \item Genderfluid
        \item Agender
        \item Bigender
        \item Other
    \end{itemize}

    \item \textbf{Please Select Your Age Group}
        \begin{itemize}
        \item less than 20
        \item 20-24
        \item 25-29
        \item 30-34
        \item 34-39
        \item 40-44
        \item 45-49
        \item 50-54
        \item 54-59
        \item 60-64
        \item 65 and older
    \end{itemize}

    \item \textbf{Please Provide Your Nationality} \\
    \textit{Short answer response}

    \item \textbf{Please Select Your Race} 
    \begin{itemize}
        \item African American or Black
        \item American Indian or Alaskan Native
        \item Asian
        \item Hispanic or Latino
        \item Native Hawaiian or Other Pacific Islander
        \item White
        \item Other
    \end{itemize}

    \item \textbf{Please Provide Your Occupation} \\
    \textit{Short answer response}

    \item \textbf{Please Provide Your Relationship Status} 
    \begin{itemize}
        \item Never Married
        \item Separated
        \item Divorced
        \item Widowed
        \item Married
    \end{itemize}

    \textbf{Section 2: Self Descriptive Response \\(Min. 500 Words)}\\

    \item \textbf{Please describe yourself.} \\
    \textit{Long answer response}
    
    \item \textbf{What are your aspirations and goals for your personal life?} \\
    \textit{Long answer response}

    \item \textbf{What are your most defining traits or qualities?} \\
    \textit{Long answer response}

    \item \textbf{Please describe your average day.} \\
    \textit{Long answer response}

    \item \textbf{What are your core values, and how do they guide your decisions?} \\
    \textit{Long answer response}

    \item \textbf{What skills do you excel at, and how do you use them?} \\
    \textit{Long answer response}
\end{enumerate}

\subsection{LLM Judge Prompts Used for Evaluation} \label{appendix:prompts}

We present the LLM instructions (prompt box below) used to generate personas using GPT-4, Gemini 1.5 Pro, and DeepSeek. The prompt is divided into four settings based on the number of sociodemographic inputs provided for creating the prompt (Race, Race/Age, Race/Age/Gender, All). Below shows the example of the setting that provides all user data to create the synthetic persona.

\begin{figure*}[t]
\begin{tcolorbox}[colback=gray!10, colframe=blue!50!black, title=Persona Generation Prompt (Setting: All Sociodemographic Information), coltitle=white, fonttitle=\bfseries]

\begin{lstlisting}[language=TeX]
In this task, pretend to be a persona using the attributes provided to be a participant in a study that is being conducted with the details below. 

Study Details:
In this study, we aim to explore how individuals express and assess their own identities through self-descriptive questions. By prompting participants with questions like "describe yourself" or "describe your day", we seek to understand the unique ways individuals portray their personal experiences and personas.

Personality Attributes:
You are a <age> year-old <race> <sex> working as a <occupation> , USA. You were born in <nationality> . You earn <income> , and your relationship status is currently <relationship> . 

Task:
Write a response to the question given below as if you are this individual and answer the questions provided below. Write a full paragraph of 5-6 sentences or more. Your post must reflect your persona strongly. 

Question: <question>
\end{lstlisting}

\end{tcolorbox}
\end{figure*}

\subsection{Frameworks of Harm} \label{appendix:harm}

To contextualize the findings in our paper, we draw from established frameworks on sociotechnical harms in language technologies, particularly those outlined by \citet{blodgett2020language}, \citet{dev2022measures}, and \citet{ghosh2024generative}. These frameworks have helped identify how generative AI systems can perpetuate or amplify structural inequities through the way they construct or simulate identity. Specifically, we focus on representational harms—harms that affect how social groups are depicted, understood, and symbolically valued—rather than allocational harms (i.e., unequal distribution of resources or opportunities).

We adopt the following six harm categories as the foundation for our analysis:

\begin{enumerate}
     \item \textbf{Stereotyping.} Refers to the reproduction of overgeneralized or essentialized beliefs about individuals based on their perceived group membership (e.g., race, gender, culture). In the context of generative AI, stereotyping often appears as repetitive, reductive patterns that frame group identities through tropes rather than nuance.
     
    \item \textbf{Disparagement.} Encompasses outputs that implicitly or explicitly diminish the value, dignity, or worth of certain groups. While often subtle, disparagement can manifest through narrative structures that assign adversity, deficiency, or marginality as default states for minoritized identities.

    \item \textbf{Dehumanization.} Occurs when generated narratives omit or deny attributes associated with shared humanity—such as agency, emotion, or relational depth. By portraying individuals as symbolic or one-dimensional, models can suppress empathy and perpetuate “othering” in more implicit ways.

    \item \textbf{Erasure.} Refers to the absence or underrepresentation of particular groups or the flattening of intra-group diversity. In generative systems, erasure may result from narrow training distributions or design choices that prioritize generic, default (often dominant group) narratives.

    \item \textbf{Exoticism.} Defined as the over-amplification of certain culturally coded features (e.g., foods, clothing, language, traditions) in ways that fetishize or aestheticize difference. Exoticism can render marginalized identities more visible yet less authentic by reducing them to spectacle.

    \item \textbf{Quality of Service.} Captures disparities in the performance or accuracy of model outputs across demographic groups. When synthetic personas vary in narrative fidelity, fluency, or stylistic quality depending on identity prompts, it signals unequal model behavior and potential downstream inequities.

\end{enumerate}

\subsection{Marked Persona Analysis: Model Breakdown} \label{appendix:markedpersona}

Table \ref{tab:marked_words_llms} show the breakdown of marked words obtained for each of the language model. The results showcase that all three models show similar behaviors of accentuating racial identities for the personas it generates. The results show a combination of all the four prompt settings used. 

\begin{table*}[]
\centering
\footnotesize
\begin{tabular}{c|c|c|c}
\textbf{Race} & \textbf{Deepseek} & \textbf{Gemini} & \textbf{GPT4o} \\ \hline
\textbf{\begin{tabular}[c]{@{}c@{}}African American\\  / Black\end{tabular}} & \begin{tabular}[c]{@{}c@{}}\textbf{black},\textit{ music, folks}, lifting, \\ \textit{community}, \textbf{african}, \textit{unapologetic}, \\ \textcolor{red}{resilience}, young, just\end{tabular} & \begin{tabular}[c]{@{}c@{}}\textbf{black}, real, \textit{hustle}, stuff, \\\textit{ mentoring, grinding, music,} \\ make, \textit{neighborhood}, \textit{community}\end{tabular} & \begin{tabular}[c]{@{}c@{}}\textbf{african}, \textit{atlanta, american,} \\ \textbf{black}, \textcolor{red}{resilience}, \textit{jazz}, young, \\ \textit{community}, \textcolor{red}{justice}, woman\end{tabular} \\
\textbf{\begin{tabular}[c]{@{}c@{}}American Indian\\  / Alaska Native\end{tabular}} & \begin{tabular}[c]{@{}c@{}}\textit{land, ancestors,} \textbf{indian}, modern, \\ \textit{elders,  generations, stories,} \\ \textit{traditions, teaching, heritage}\end{tabular} & \begin{tabular}[c]{@{}c@{}}\textit{land, heritage, community,}\\ \textit{traditions, ancestors, reservation,} \\ \textit{respect, elders, stories, tribal}\end{tabular} & \begin{tabular}[c]{@{}c@{}}\textit{ancestors,} \textbf{indian}, \textit{generation},\\  \textit{tradition, heritage, wisdom,} \\ \textit{passed, land, member, earth}\end{tabular} \\
\textbf{Asian} & \begin{tabular}[c]{@{}c@{}}\textbf{asian}, \textit{culture}, \textbf{american}, \\ \textit{immigrant, parent, heritage,} \\ \textit{identity, expectations, duality,} \\ living\end{tabular} & \begin{tabular}[c]{@{}c@{}}\textit{parent,} \textbf{korean}, \textit{first\_generation}, \\ \textit{introverted, kdrama,} single, \\ \textit{boba,} \textbf{american}, \textit{preop, vietnam}\end{tabular} & \begin{tabular}[c]{@{}c@{}}\textbf{asian}, \textit{culture, sanfrancisco,} \\ \textit{heritage, diverse, fusion,} \\ \textit{blend, roots, diversity, usa}\end{tabular} \\
\textbf{\begin{tabular}[c]{@{}c@{}}Hispanic\\  / Latino\end{tabular}} & \begin{tabular}[c]{@{}c@{}}\textbf{latino}, \textbf{hispanic}, \textit{spanish}, \\ \textit{con, abuela, parents, roots,} \\ \textit{family, abuela, mexico,}\end{tabular} & \begin{tabular}[c]{@{}c@{}}\textit{mexico, abuela, latina,} \\ \textit{maria, dios, family, miguel,} \\ \textit{sofia, family, roots}\end{tabular} & \begin{tabular}[c]{@{}c@{}}\textbf{hispanic}, \textbf{latino}, \textit{cultural}, \\ h\textit{eritage, vibrant, latina,} \\ \textit{family, spanish, abuela, roots}\end{tabular} \\
\textbf{White} & \begin{tabular}[c]{@{}c@{}}\textbf{white}, \textit{usa}, pretty, say, \\ work, think, person, project, \\ friends, weekend\end{tabular} & \begin{tabular}[c]{@{}c@{}}\textit{sarah}, husband, well, \\ director, bless, weekend, \\ \textit{ohio,} coffee, bakery, honesty\end{tabular} & \begin{tabular}[c]{@{}c@{}}\textit{midwest}, town, \textit{oregon},\\ \textit{portland}, \textbf{white}, small, \\ \textit{suburb}, years, \textit{hiking}, coffee\end{tabular}
\end{tabular}
\caption{
Top 10 marked words in the 3 LLM-generated personas (Deepseek, Gemini and GPT4o) for each racial group, identified using log-odds ratio analysis with White personas as the reference group. Words in \textbf{bold} directly denote racial or ethnic identity, \textit{italicized} terms reflect culturally coded or identity-relevant associations, and words in \textcolor{red}{red} indicate adversity-oriented language associated with \textit{trauma scripting}.
}
\label{tab:marked_words_llms}
\end{table*}

\subsection{Marked Persona Analysis: Prompt Settings Breakdown} \label{appendix:markedpersona_setting}

Table \ref{tab:marked_words_setting} show the breakdown of marked words obtained for all the language models together for the four prompts settings used in creating each of the personas. Each setting represented the number of sociodemographic inputs provided to construct the persona. Our results showed that irrespective of the setting used, the model tends to fixate on the markedness of the persona with respect to their racial features.  

\begin{table*}[]
\centering
\subfootnotesize
\begin{tabular}{c|c|c|c|c}
\textbf{} & \textbf{Race} & \textbf{Race/Age} & \textbf{Race/Age/Gender} & \textbf{All} \\ \hline
\textbf{\begin{tabular}[c]{@{}c@{}}African American\\  / Black\end{tabular}} & \begin{tabular}[c]{@{}c@{}}\textbf{black, african,} \textcolor{red}{resilience}, \\ woman, \textit{atlanta}, \textcolor{red}{strength}, \\ \textit{music, folk,} \textcolor{red}{justice}, \\ \textit{community}\end{tabular} & \begin{tabular}[c]{@{}c@{}}\textbf{african, black}, \textit{atlanta}, \\ \textit{community}, \textcolor{red}{resilience}, woman, \\ music, \textbf{american}, \textcolor{red}{justice}, \\ young\end{tabular} & \begin{tabular}[c]{@{}c@{}}\textbf{black, african,} \textit{atlanta},\\  \textit{community}, man, \\ \textcolor{red}{resilience}, music, \\ young, \textcolor{red}{lifting, uplift}\end{tabular} & \begin{tabular}[c]{@{}c@{}}\textbf{black, african}, man, \\ \textit{community, hustle,} real, \\ young, music,\\ \textcolor{red}{grinding, resilience}\end{tabular} \\
\textbf{\begin{tabular}[c]{@{}c@{}}American Indian\\  / Alaska Native\end{tabular}} & \begin{tabular}[c]{@{}c@{}}\textit{our, land, ancestors,} \\ \textit{generation, traditions,} \\ \textbf{indian}, \textit{stories,  teaching,} \\ \textit{passed, earth}\end{tabular} & \begin{tabular}[c]{@{}c@{}}\textit{ancestor, generation, land, }\\ \textit{stories, traditions, passed, }\\ \textit{elders,  heritage,} modern, \\ \textit{wisdom}\end{tabular} & \begin{tabular}[c]{@{}c@{}}our, \textit{ancestor, land}, \\ \textit{generation, traditions,} \\ passed, \textit{stories, elders,}\\ \textbf{indian}, wisdom\end{tabular} & \begin{tabular}[c]{@{}c@{}}\textbf{indian}, \textit{heritage}, \\ \textit{community, traditions,} \\ \textit{ancestor, land, respect}, \\ \textit{elders, passed,} \textbf{american}\end{tabular} \\
\textbf{Asian} & \begin{tabular}[c]{@{}c@{}}\textbf{asian}, \textit{parents, culture,} \\ \textbf{american}, \textit{heritage, blend}, \\ \textit{balancing, navigating,} \\ \textit{diverse, worlds}\end{tabular} & \begin{tabular}[c]{@{}c@{}}\textbf{asian}, \textit{culture, parents}, \\ \textit{heritage,} \textbf{american}, \\ \textit{expectation}, \\ \textbf{korean}, \textit{immigrant}, \\ \textit{diverse, kimchi}\end{tabular} & \begin{tabular}[c]{@{}c@{}}\textbf{asian}, \textit{culture}, \\ \textit{parents},  \textbf{korean}, \textit{heritage}, \\ \textit{sanfrancisco, diverse,} \\\textit{ software, engineer, kimch}i\end{tabular} & \begin{tabular}[c]{@{}c@{}}\textbf{asian}, \textit{parent, cultural}, \\ \textbf{american}, \textbf{korean}, \textit{hr}, \\ \textit{firstgen, students},\\ physicians, test\end{tabular} \\
\textbf{\begin{tabular}[c]{@{}c@{}}Hispanic\\  / Latino\end{tabular}} & \begin{tabular}[c]{@{}c@{}}\textbf{latino}, \textit{abuela, con},\\  \textit{spanish, hispanic, salsa,} \\ \textit{family, arroz. tamales,}\\  proud\end{tabular} & \begin{tabular}[c]{@{}c@{}}latino, abuela, hispanic, \\ con, spanish, cultural,\\ \textit{family, arroz, salsa,} \\ \textit{roots}\end{tabular} & \begin{tabular}[c]{@{}c@{}}\textbf{latino, hispanic, latina,} \\ \textit{abuela, con, family,} \\ \textit{spanish, cultural},\\ \textit{salsa, leche}\end{tabular} & \begin{tabular}[c]{@{}c@{}}\textbf{hispanic, latina, mexico}, \\ \textbf{latino}, \textit{roots}, \textbf{spanish}, \\ \textit{family, cultural},\\ \textit{abuela, vibrant}\end{tabular} \\
\textbf{White} & \begin{tabular}[c]{@{}c@{}}pretty, \textbf{white}, honesty, \\ \textit{midwest}, town, try, \\ \textit{hiking}, guy, persona, \\ straightforward\end{tabular} & \begin{tabular}[c]{@{}c@{}}\textbf{white}, pretty, honesty, \\ dog, \textit{hiking}, team, \\ town, years,\\ coffee, marketing\end{tabular} & \begin{tabular}[c]{@{}c@{}}\textbf{white}, honesty, coffee, \\ \textit{beer, midwest, hiking,} \\ well, marketing,\\ town, league\end{tabular} & \begin{tabular}[c]{@{}c@{}}husband, bakery, \textbf{white}, \\ clients, years, business, \\ advisor, director,\\ editor, weekend\end{tabular}
\end{tabular}
\caption{
Top 10 marked words in the 4 prompt setting used for all the 3 LLMs combined, for each racial group, identified using log-odds ratio analysis with White personas as the reference group. Words in \textbf{bold} directly denote racial or ethnic identity, \textit{italicized} terms reflect culturally coded or identity-relevant associations, and words in \textcolor{red}{red} indicate adversity-oriented language associated with \textit{trauma scripting}.
}
\label{tab:marked_words_setting}
\end{table*}

\end{document}